\documentclass[11pt]{article}

\usepackage[final]{acl}

\usepackage{times}
\usepackage{latexsym}

\usepackage[T1]{fontenc}

\usepackage{times}
\usepackage{latexsym}
\usepackage[T1]{fontenc}
\usepackage[utf8]{inputenc}
\usepackage{microtype}
\usepackage{inconsolata}
\usepackage{graphicx}
\usepackage{booktabs}
\usepackage{amsmath}
\usepackage{amssymb}
\usepackage{pifont}
\newcommand{\cmark}{\ding{51}}%
\newcommand{\xmark}{\ding{55}}%
\usepackage{multirow}
\usepackage{xcolor}
\usepackage{caption}
\usepackage{subcaption}
\usepackage{xspace}
\usepackage{algorithm}
\usepackage{algpseudocode}
\usepackage{tikz}
\usetikzlibrary{shapes.geometric, arrows.meta, positioning, fit, backgrounds}
\usepackage{enumitem}
\usepackage{placeins}
\usepackage{bbm}
\usepackage{tcolorbox}
\tcbuselibrary{breakable,skins}
\tcbset{promptstyle/.style={%
  enhanced jigsaw, breakable, pad at break*=1mm, colback=lightgray, colframe=flatblue,
  boxrule=0.6pt, arc=3pt, left=8pt, right=8pt, top=5pt, bottom=5pt
}}

\definecolor{flatblue}{HTML}{1f77b4}
\definecolor{hierred}{HTML}{d62728}
\definecolor{loragreen}{HTML}{2ca02c}
\definecolor{lightgray}{HTML}{F5F5F5}

\newcommand{\toolgen}{ToolGen\xspace}

\usepackage{microtype}

\usepackage{inconsolata}

\usepackage{graphicx}
\usepackage{url}

\title{ToolSense: A Diagnostic Framework for Auditing\\Parametric Tool Knowledge in LLMs}

\author{
  Ashutosh Hathidara\thanks{Equal contribution.} \quad
  Sai Shruthi Sistla\footnotemark[1] \quad
  Sebastian Schreiber \quad
  Sahil Bansal \\
  SAP Labs \\
  {\small \texttt{\{ashutosh.hathidara, sai.shruthi.sistla, sebastian.schreiber, sahil.bansal01\}@sap.com}}
}

\begin{document}
\maketitle
\begin{abstract}
Large language models deployed as agents over large tool catalogs face a critical
tool-retrieval bottleneck. As embedding-based retrieval approaches rely on compact
encoders that may under-capture specialized tool semantics, parametric tool
retrieval~\citep{qin2025toolgen} addresses this by encoding each tool as a
\emph{virtual token} appended to the LLM vocabulary, fine-tuned in two stages
(memorization then retrieval SFT) to use the LLM as a retriever, achieving strong
performance on standard ToolBench retrieval benchmarks. Yet these benchmarks use verbose,
fully-specified queries, and their evaluation applies constrained decoding that
restricts outputs to valid token paths --- neither reveals whether the model actually
\emph{understands} its tools. We introduce \textbf{ToolSense}, an open-source
LLM-powered diagnostic framework that takes any tool catalog as input and automatically
generates three benchmarks: a Realistic Retrieval Benchmark~(RRB) with queries at
three ambiguity tiers, an MCQ probing benchmark, and a QA probing benchmark.
Applying ToolSense to ToolBench~($\sim$47k tools) and evaluating five parametric model training
configurations reveals a \emph{knowledge-retrieval dissociation}: on RRB queries,
several configurations collapse by \textbf{$\sim$50-64 percentage points} compared to
fully-specified ToolBench benchmarks, falling below embedding-model baseline. Additionally, despite
strong retrieval performance, some models score near-random on factual
probes, suggesting a knowledge-retrieval dissociation. We open-source the ToolSense
framework and the ToolBench diagnostic benchmarks at \url{https://github.com/SAP/toolsense}.
\end{abstract}

\section{Introduction}
\label{sec:intro}

Large language models are increasingly deployed as autonomous agents over large software ecosystems,
where selecting the right tool from catalogs of thousands of APIs is a core bottleneck
\citep{schick2023toolformer,yao2023react,qu2025toolsurvey}.
The dominant approach encodes tool descriptions as dense vectors and retrieves top-$k$ candidates
via approximate nearest-neighbor search~\citep{karpukhin2020dpr}.
This paradigm has known limitations: small retrieval encoders may underspecify the semantics of
complex, overlapping APIs at scale~\citep{qin2025toolgen}; retrieved descriptions must still be injected into the LLM
context, compounding overhead across multi-step agent loops; and retriever and generator are
trained with separate objectives, preventing joint optimization toward task success~\citep{tay2022dsi}.

\textbf{Parametric Tool Retrieval.}
\citet{qin2025toolgen} propose ToolGen, an alternative strategy to encode the entire tool catalog directly into LLM
parameters. Each tool is assigned a unique \emph{virtual token} appended to the model vocabulary as a flat atomic identifier representing the full tool identity (e.g., \texttt{<<TIKTOK\&\&GET\_TRENDING\_VIDEOS>>}). In Stage~1 (memorization), the model is fine-tuned to associate
each tool's metadata with its virtual token. In Stage~2 (retrieval), the model is further
fine-tuned on (query $\to$ virtual token) pairs, learning to generate the correct token given a
natural language query. At inference, a \emph{DisjunctiveTrie}~\citep{decao2021genre} constrains beam search
to valid token sequences, guaranteeing every output maps to a real tool. Applied to
ToolBench~\citep{qin2023toolllm} ($\sim$47k tools), this paradigm achieves over $\approx 0.90$ recall on standard ToolBench benchmarks.

\textbf{The Diagnostic Gap.}
This naturally raises a question: \emph{does parametric training produce a model that understands
its tools, or does it simply learn to match query patterns to token identifiers?}
A model that has learned only surface-level mappings may appear to work on distribution-matched
benchmarks while failing the moment a user phrases their query differently.
Yet existing evaluation is not designed to reveal this, for two reasons.
First, the queries in ToolBench's standard evaluation splits (G1/G2/G3) are verbose
and fully specified, unlike how a real user would phrase them (examples in Appendix~\ref{sec:appendix-query-comparison}).
Second, constrained decoding means the model need only rank paths through a fixed trie
rather than freely recall a tool token, masking whether genuine memorization occurred.
This matters practically: downstream agentic fine-tuning typically requires free-form
token generation without trie support, making undetected memorization gaps a silent
failure risk.
Together, verbose queries and constrained decoding make it difficult to distinguish
pattern matching from genuine tool knowledge~\citep{petroni2019lama}, motivating a
more diagnostic evaluation protocol.

We introduce \textbf{ToolSense}, an open-source LLM-powered diagnostic framework that takes any
tool catalog as input and automatically generates three diagnostic benchmarks:
(1)~a \emph{Realistic Retrieval Benchmark}~(RRB) with queries in three ambiguity tiers reflecting how human users
actually write queries;
(2)~an \emph{MCQ probing benchmark} testing discriminative factual knowledge about tool
capabilities; and
(3)~a \emph{QA probing benchmark} testing inferential factual knowledge about tool properties.
Alongside the benchmarks, we establish a free-form evaluation protocol that reports an
\emph{Internalization Score} (IS@k = free@k / constrained@k) as a trie-dependency diagnostic.

Applying ToolSense to ToolBench reveals a striking gap: configurations that appear
highly capable on standard benchmarks collapse dramatically on realistic queries,
falling below non-parametric baselines. Factual probing further reveals that Stage~2 training (responsible for the retrieval gains) nearly universally destroys the tool knowledge acquired in Stage~1.

To summarize, our contributions include: \textbf{(i)}~\textbf{ToolSense}, an open-source
diagnostic framework that auto-generates RRB, MCQ, and QA benchmarks from any tool
catalog; \textbf{(ii)}~\textbf{ToolBench diagnostic benchmarks} RRB, MCQ, and QA released openly; and \textbf{(iii)}~an empirical diagnosis
of five parametric configurations on ToolBench, revealing the training choices that govern the
knowledge-retrieval dissociation.

\section{Background and Related Work}
\label{sec:related}

\textbf{Generative and parametric retrieval.}
\citet{tay2022dsi} introduced Differentiable Search Index~(DSI), encoding document IDs into
transformer parameters and retrieving via constrained beam search.
\citet{decao2021genre} applied autoregressive entity retrieval to knowledge-grounded generation.
\citet{wang2022nci} extended this to dense document corpora.
\citet{qin2025toolgen} adapt the paradigm to tool catalogs with virtual tokens and two-stage
training. All of these systems evaluate exclusively with constrained decoding; our free-form IS
protocol introduces a new diagnostic practice for this entire class of systems.

\textbf{Tool learning in LLMs.}
\citet{qin2023toolllm} build ToolBench, a benchmark covering $\sim$16k real-world APIs ($\sim$47k tools).
\citet{patil2023gorilla} fine-tune LLMs to generate tool-calls with retrieval augmentation.
\citet{schick2023toolformer} train LLMs to self-insert tool-calls in context.
We focus specifically on the \emph{retrieval} stage of parametric tool systems and ask whether
that retrieval mechanism encodes tool semantics, a question orthogonal to downstream task
performance studied by these works.


\textbf{Probing and interpretability.}
\citet{petroni2019lama} introduce LAMA, probing pre-trained LMs for factual knowledge via
cloze-style queries. \citet{tenney2019bert} probe BERT's representations for linguistic
structure. Our MCQ and QA probes extend this tradition to a novel setting: tokens
\emph{learned during fine-tuning}~(virtual tokens), not pre-trained representations.
We probe whether fine-tuning creates genuinely semantic representations or only task-relevant pointers.

\textbf{Knowledge retention in fine-tuning.}
\citet{mccloskey1989catastrophic} identify catastrophic interference as a fundamental
failure mode when neural networks are trained sequentially on new tasks.
\citet{hu2022lora} propose LoRA to mitigate this by freezing backbone weights and
learning low-rank updates, preserving pre-trained representations.
\citet{biderman2024lora} empirically show that LoRA retains more prior task performance
than full fine-tuning. Our approach extends this line of work to the parametric
retrieval setting.

\section{The ToolSense Framework}
\label{sec:framework}

\begin{figure*}[t]
\centering
\includegraphics[width=0.80\linewidth, trim = 0.5cm 0cm 0.5cm 0cm, clip]{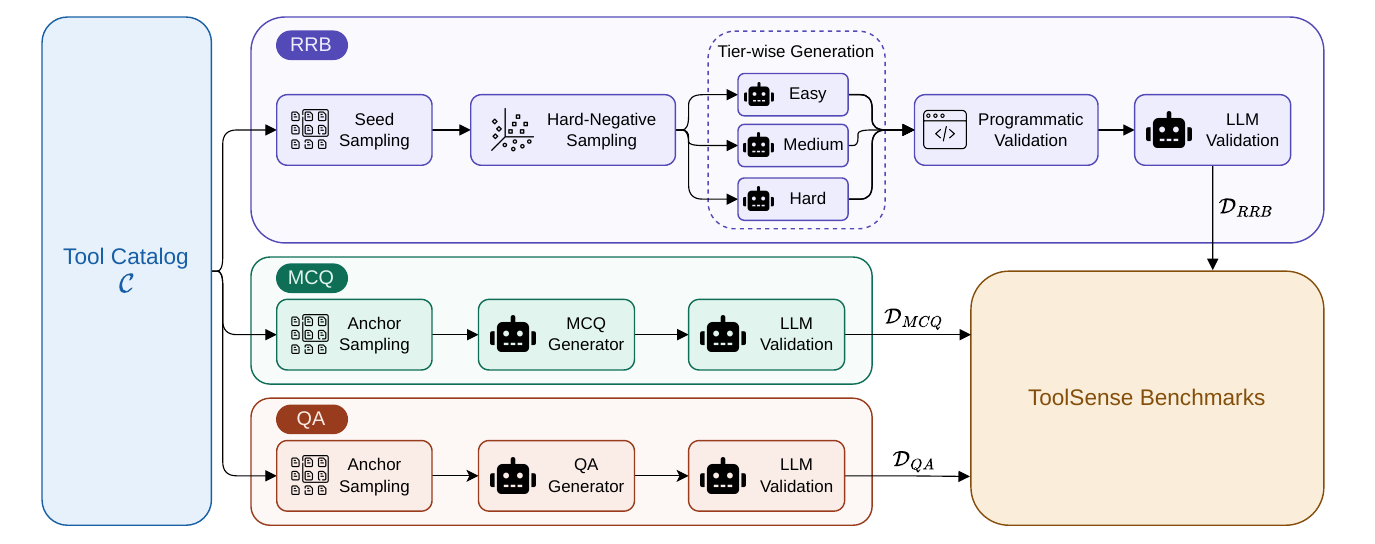}
\caption{The ToolSense diagnostic framework generates $\mathcal{D}_\mathrm{RRB}$, $\mathcal{D}_\mathrm{MCQ}$, and $\mathcal{D}_\mathrm{QA}$ from a tool catalog $\mathcal{C}$.}
\label{fig:framework}
\end{figure*}

The framework takes as input a \textbf{tool catalog}
\begin{equation}
  \mathcal{C} = \bigl\{\,\tau_i = (\mathrm{name}_i,\; \mathrm{desc}_i)\,\bigr\}_{i=1}^{|\mathcal{C}|}
  \label{eq:catalog}
\end{equation}
and automatically generates three diagnostic benchmark datasets.
Figure~\ref{fig:framework} illustrates the pipeline. In the following subsections, we explain the benchmark generation methodology in detail.

\subsection{Realistic Retrieval Benchmark (RRB)}
\label{sec:rrb}

RRB tests whether a retrieval system generalises beyond verbose-style queries by
presenting it with short, intent-focused requests at three ambiguity levels.
Generation proceeds in four stages: seed sampling, hard-negative pool construction,
parallel generation tiers, and dual validation.

\textbf{Stratified seed sampling.}
Rather than generating queries for every tool in $\mathcal{C}$, we sample
$N_{\mathrm{RRB}}$ anchor tools stratified by domain to ensure broad catalog
coverage, yielding $\mathcal{C}_{\mathrm{RRB}} \subset \mathcal{C}$ with
$|\mathcal{C}_{\mathrm{RRB}}| = N_{\mathrm{RRB}}$.

\textbf{Hard-negative pool.}
For each anchor $\tau \in \mathcal{C}_{\mathrm{RRB}}$, let $\phi: \mathcal{C} \to \mathbb{R}^d$
be a sentence encoder. We retrieve the $K$ nearest neighbours by cosine
similarity, $\mathcal{H}_K(\tau) = \arg\operatorname{top-}K_{\tau' \in \mathcal{C} \setminus \{\tau\}}
\langle \phi(\tau), \phi(\tau') \rangle$, and form the candidate pool
$\mathcal{P}(\tau) = \{\tau\} \cup \mathcal{H}_K(\tau)$ with $|\mathcal{P}(\tau)| = K{+}1$.
The pool defines a hard-negative context shown to the generator and the labels from which ground-truth answers $A$ are drawn.

\textbf{Parallel generation tiers.}
For each anchor, three sub-pipelines run in parallel,
$t \in \{\mathrm{easy},\, \mathrm{medium},\, \mathrm{hard}\}$, each generating a batch
of $(q, A)$ pairs from the same candidate pool $\mathcal{P}(\tau)$:
\begin{itemize}[topsep=2pt,itemsep=1pt]
  \item \textbf{Easy}: $|A|{=}1$; a concise query pointing to exactly one tool.
  \item \textbf{Medium}: $|A| \in \{2,3\}$; a cross-functional request genuinely
        satisfiable by 2--3 tools.
  \item \textbf{Hard}: $|A|{\geq}4$; a high-level, ambiguous goal spanning 4 or more tools.
\end{itemize}
The generator LLM $\theta_{\mathrm{RRB}}$ receives $(\tau,\, \mathcal{P}(\tau),\, t,\, \mathcal{E})$,
where $\mathcal{E} = \{\hat{q}_i^{(e)}\}_{i=1}^{n_e}$ are few-shot query examples provided to
guide the style and tone of intent-focused query generation.

\textbf{Dual validation with feedback.}
Each generated batch passes two sequential filters. The programmatic filter $\mathcal{V}_{RRB}(q, A, \tau, \mathcal{P}(\tau))$
rejects entries failing either of two checks: answers must be grounded in the pool
($A \subseteq \mathcal{P}(\tau)$), and no answer tool name may appear verbatim in the query
($\forall\,\tau' \in A : \mathrm{name}(\tau') \notin q$).
Entries passing $\mathcal{V}_{RRB}$ are then scored by an LLM judge $\mathcal{J}_{RRB}$ for naturalness,
tier-compliance, and label correctness.
Rejected entries are dropped, yielding the validated dataset $\mathcal{D}_{\mathrm{RRB}} = \{(q_j,\, A_j,\, t_j)\}_{j=1}^{N_{\mathrm{RRB}}}$.


\subsection{Probing Benchmarks (MCQ and QA)}
\label{sec:probes}

Both probing benchmarks follow the same pipeline structure: sample a seed set of tools, generate question--answer pairs via an LLM generator, validate with an LLM judge $\mathcal{J}$, and filter entries for which no unambiguous question can be formed.
In all accepted entries, the question refers to the tool only as ``this tool'', never by name, so the model must answer given the virtual token $v_\tau$.

\textbf{MCQ} ($N_{\mathrm{MCQ}}$ tools).
For each $\tau$, $\theta_{\mathrm{MCQ}}$ produces a factual question $z$ about the tool's functionality, one correct answer $o_{y^\star}$, and three plausible-but-wrong distractors, yielding $\mathcal{D}_{\mathrm{MCQ}} = \{(v_{\tau_i},\,z_i,\,\{o_k^i\},\,y_i^\star)\}$.
At evaluation the model receives $(v_\tau,\,z,\,\{o_k\}_{k=1}^{4})$ and is scored by \emph{constrained first-token probability}: we read the model's next-token log-probability distribution, restrict it to the closed candidate set of option letters (A/B/C/D), and take the highest-probability option as the prediction $\hat{y}$, which is compared with $y^\star \in \{0,1,2,3\}$.

\textbf{QA} ($N_{\mathrm{QA}}$ tools).
For each $\tau$, $\theta_{\mathrm{QA}}$ receives the description alongside a pre-specified target $b \in \{\mathrm{Yes},\mathrm{No}\}$, alternated across tools to ensure label balance, and produces a binary question $w$ about a specific, verifiable tool property (e.g., supported modality, domain, etc.), yielding $\mathcal{D}_{\mathrm{QA}} = \{(v_{\tau_i},\,w_i,\,b_i)\}$.
At evaluation the model receives $(v_\tau,\,w)$ and is scored analogously by first-token probability over the closed set $\{\mathrm{Yes},\mathrm{No}\}$, comparing $\hat{b}$ with $b$.

Because both probes read directly off the model's probability distribution over a fixed candidate-token set rather than requiring the model to freely generate a letter or word, they are format-agnostic: a model that has internalized tool semantics scores highly regardless of its preferred output surface form, whereas free-generation scoring would conflate knowledge with output-format compliance. This constrained-token scoring is the standard methodology in LLM knowledge probing~\citep{petroni2019lama,3495724.3495883,hendrycks2021measuring}.

\subsection{Free-form Evaluation and Internalization Score}
\label{sec:is}

Let $\hat{A}_c(q, k)$ and $\hat{A}_f(q, k)$ denote the top-$k$ outputs under
\emph{constrained} (trie-guided) and \emph{free-form}
(unconstrained) decoding respectively using beam search with beam width $B$.
Constrained and free-form Recall@$k$ over query set $\mathcal{Q}$ are defined as
$R_c@k = \frac{1}{|\mathcal{Q}|}\sum_{q \in \mathcal{Q}} \mathbbm{1}[A(q) \cap \hat{A}_c(q,k) \neq \emptyset]$
and $R_f@k = \frac{1}{|\mathcal{Q}|}\sum_{q \in \mathcal{Q}} \mathbbm{1}[A(q) \cap \hat{A}_f(q,k) \neq \emptyset]$.
The \textbf{Internalization Score} (IS) is then:
\begin{equation}
  \text{IS@k} \;=\; \begin{cases} R_f@k \;/\; R_c@k & \text{if } R_c@k > 0 \\ 0 & \text{otherwise} \end{cases}
  \label{eq:is}
\end{equation}
IS $\approx 1$ indicates the model generates correct tokens equally well without the trie
(retrieval is internalized); IS $\approx 0$ indicates complete trie-dependence.
IS is a ratio of two standard measurements, which enables us to \emph{always} report $R_f@k$ alongside $R_c@k$, making trie-dependence transparent. (see Appendix~\ref{sec:appendix-is-rationale} for design rationale for IS.)


\section{Experimental Setup}
\label{sec:experimental-setup}

We instantiate ToolSense on the ToolBench~\citep{qin2023toolllm} tool catalog to empirically diagnose parametric tool retrieval across three dimensions: OOD generalization beyond training-distribution queries, dependence on constrained decoding, and correspondence between retrieval performance and semantic tool knowledge.
All trained models follow the two-stage \toolgen paradigm~\citep{qin2025toolgen}: Stage~1 (memorization) fine-tunes on (tool metadata $\rightarrow$ virtual token) pairs; Stage~2 (retrieval) fine-tunes on (query $\rightarrow$ virtual token) pairs.





\paragraph{Training and evaluation data.}
Stage~1 uses 46,980 RapidAPI tools provided by ToolBench, each with a api name \& description as well as endpoint name \& description. Stage~2 uses $\sim$195k verbose (query, tool) pairs from the ToolBench training split~\citep{qin2023toolllm}, generated by GPT-4~\citep{openai2024gpt4technicalreport}.
For standard evaluation, we retain the \toolgen evaluation splits: G1 (593 queries), G2 (399), and G3 (100)~\citep{qin2025toolgen}; these match the verbose training distribution and serve as the in-distribution baseline (see Appendix~\ref{sec:appendix-query-comparison} for a query style comparison).

\paragraph{Diagnostic benchmarks.}
Applying the ToolSense generation pipelines (\S\ref{sec:rrb}--\S\ref{sec:probes}) to this catalog yields: an RRB of 500 queries across three difficulty tiers (167 easy / 167 medium / 166 hard), an MCQ probe of 496 samples testing discriminative factual knowledge, and a QA probe of 500 items testing inferential knowledge. Together, these evaluate retrieval generalization (RRB) and semantic understanding (MCQ/QA) as orthogonal dimensions of tool knowledge.

\subsection{Model Configurations}
\label{sec:model-configs}

We primarily use open-source instruction-tuned \textbf{Gemma3-4B}~\citep{google2025gemma3}, with \textbf{Qwen3.5-4B}~\citep{qwen2025qwen3} for cross-architecture validation and \textbf{Gemma3-12B} for scale ablations.
For each configuration we train both a full fine-tuned~(FFT) and a LoRA~\citep{hu2022lora} variant; training details are in Appendix~\ref{sec:appendix-hyperparams}. In all LoRA variants, adapters are attached to linear layers only, while the full embedding layer remains trainable.

We define five primary training configurations in Table~\ref{tab:configs}, each isolating one variable along three axes: \emph{token format} (flat single-token vs.\ hierarchical multi-token), \emph{memorization format} (single-format vs.\ multi-format with reverse mappings and hard negatives), and \emph{training method} (FFT vs.\ LoRA). Flat tokens follow the original \toolgen format (\S\ref{sec:intro}); hierarchical tokens decompose the identifier into a two-step sequence \texttt{<<API>><<Endpoint>>} (e.g., \texttt{<<TIKTOK>><<GET\_TRENDING\_VIDEOS>>}) that the model should generate autoregressively.

\textbf{TG} replicates the original \toolgen setup (flat tokens, desc$\to$tok only, no system prompt) and serves as the baseline. \textbf{TG-SP} adds a task-specific system prompt applied consistently across both training stages and inference. \textbf{TG-3FM} extends TG-SP with two additional Stage~1 formats (tok$\to$desc and Multi-Choice Tool Selection~(MCTS, Appendix~\ref{sec:appendix-mcts})), testing whether richer memorization objectives produce more robust representations. \textbf{TG-H} switches from flat to hierarchical tokens while keeping all else matched to TG-SP, isolating the effect of token structure. \textbf{TG-5FM} combines hierarchical tokens with all five memorization formats (Table~\ref{tab:configs}), where the additional formats train each hierarchical generation step independently. Where applicable, we additionally train a \textbf{LoRA} variant for the above configurations. Full training details for each configuration are in Appendix~\ref{sec:appendix-configs}.

Stage~2 training uses identical (query $\to$ tok) data across all configurations; the only variation is whether a system prompt is prepended, consistent with how each configuration was trained in Stage~1.

\begin{table}[t]
  \centering
  \small
  \resizebox{\columnwidth}{!}{%
  \begin{tabular}{@{}lllcc@{}}
    \toprule
    Config & Tokens & Tool Memo. Formats & SP & Train \\
    \midrule
    TG        & Flat & 1 (desc$\rightarrow$tok) & \xmark & FFT, LoRA \\
    TG-SP     & Flat & 1 (desc$\rightarrow$tok) & \cmark & FFT, LoRA \\
    TG-3FM    & Flat & 3 (desc$\rightarrow$tok, tok$\rightarrow$desc, MCTS) & \cmark & FFT, LoRA \\
    TG-H      & Hier. & 1 (desc$\rightarrow$tok) & \cmark & FFT \\
    TG-5FM    & Hier. & 5 (all$^*$) & \cmark & FFT, LoRA \\
    \bottomrule
    \multicolumn{5}{@{}l}{\footnotesize $^*$desc$\rightarrow$tok, desc$\rightarrow$api\_tok, desc+api\_tok$\rightarrow$endpoint\_tok, tok$\rightarrow$desc, MCTS}
  \end{tabular}}
  \caption{Diagnostic model configurations. \texttt{tok} denotes the tool token encoding the full tool identity; \texttt{api\_tok} and \texttt{endpoint\_tok} are the two hierarchical tokens corresponding to API and endpoint respectively. SP = system prompt.}
  \label{tab:configs}
\end{table}

\paragraph{Retrieval baselines.}
We compare against \textbf{BM25}~\citep{Robertson2009bm25} (sparse lexical) and \textbf{text-embedding-3-large}~\citep{OpenAI2024te3l} (te3l; OpenAI dense embeddings) on all evaluation splits.

\paragraph{Evaluation metrics.}

$R_c@k$, $R_f@k$, and IS@$k$ follow the formal definitions in \S\ref{sec:is} and are reported at beam width $B{=}50$ unless otherwise specified.
MCQ and QA probing benchmarks are evaluated by accuracy under the constrained first-token probability scoring defined in \S\ref{sec:probes}: the prediction is the highest-probability token in the closed candidate set (A/B/C/D for MCQ, Yes/No for QA); random baselines are 25\% and 50\%, respectively.

\begin{figure}[t]
  \centering
  \begin{subfigure}{0.48\linewidth}
    \includegraphics[width=\linewidth]{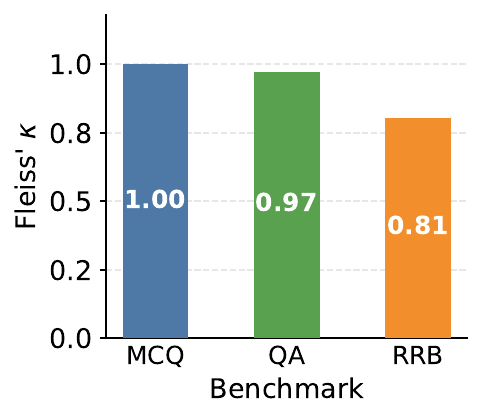}
    \caption{Inter-annotator agreement Fleiss' $\kappa$ per benchmark.}
    \label{fig:human-kappa}
  \end{subfigure}
  \hfill
  \begin{subfigure}{0.48\linewidth}
    \includegraphics[width=\linewidth]{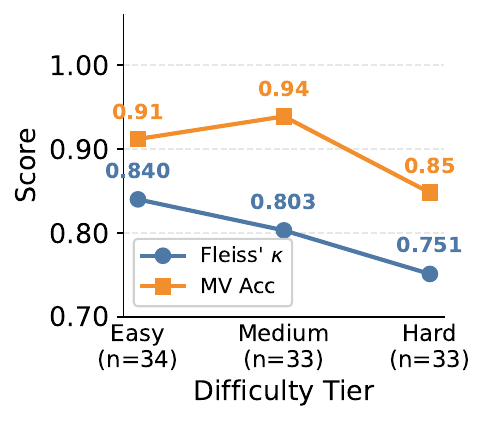}
    \caption{RRB breakdown by difficulty tier.}
    \label{fig:human-rrb-tiers}
  \end{subfigure}
  \caption{Human annotation study ($N=100$ items $\times$ 3 annotators). RRB agreement decreases gracefully with difficulty while majority-vote accuracy remains high (85--94\%).}
  \label{fig:human-study}
\end{figure}

\paragraph{Benchmark quality validation.}
Three human annotators independently validated a stratified 100 samples from each benchmark: for RRB, annotators labeled from 14 candidate tools (ground truths + hard negatives) that matched each query's intent; for MCQ/QA, annotators answered each item given the full tool description.
Figure~\ref{fig:human-kappa} shows Fleiss' $\kappa$ across all three benchmarks: MCQ achieves perfect agreement ($\kappa = 1.000$), QA near-perfect ($\kappa = 0.973$), and RRB substantial agreement ($\kappa = 0.805$).
Figure~\ref{fig:human-rrb-tiers} breaks down RRB agreement by difficulty tier: $\kappa$ decreases gracefully from easy (0.840) to hard (0.751) while remaining above agreement threshold (0.70), confirming that the tier structure reflects genuine task difficulty rather than benchmark artifacts.

\section{Results}
\label{sec:results}

We present findings across retrieval generalization, constrained decoding dependence, and semantic tool knowledge.

\subsection{Generalization Collapse on Realistic Queries}
\label{sec:results-rrb}

\begin{table*}[t]
  \centering
  \small
  \resizebox{\textwidth}{!}{%
  \begin{tabular}{@{}l cc cc cc cc r@{}}
    \toprule
    & \multicolumn{2}{c}{G1 (\%)} & \multicolumn{2}{c}{G2 (\%)} & \multicolumn{2}{c}{G3 (\%)} & \multicolumn{2}{c}{RRB (\%)} & \\
    \cmidrule(lr){2-3} \cmidrule(lr){4-5} \cmidrule(lr){6-7} \cmidrule(lr){8-9}
    Config & S1 & S2 & S1 & S2 & S1 & S2 & S1 & S2 & $\Delta$pp \\
    \midrule
    TG & \underline{47.0\,{\tiny[43.6,50.4]}} & \textbf{95.7\,{\tiny[94.2,97.0]}} & \underline{37.0\,{\tiny[33.8,40.4]}} & 84.5\,{\tiny[81.5,87.3]} & \textbf{35.0\,{\tiny[27.9,42.7]}} & 84.6\,{\tiny[80.4,88.5]} & \underline{37.8\,{\tiny[34.3,41.2]}} & 43.8\,{\tiny[40.3,47.7]} & $-$52 \\
    TG (LoRA) & 38.8\,{\tiny[35.2,42.4]} & 90.4\,{\tiny[88.7,92.2]} & 28.4\,{\tiny[25.3,31.6]} & 82.8\,{\tiny[79.8,85.5]} & 26.8\,{\tiny[20.0,34.0]} & 84.9\,{\tiny[79.9,89.6]} & 31.8\,{\tiny[28.2,35.2]} & 37.0\,{\tiny[33.4,40.7]} & $-$53 \\
    TG-SP & 38.6\,{\tiny[35.0,42.4]} & 94.5\,{\tiny[92.9,96.0]} & 28.5\,{\tiny[25.4,31.5]} & \underline{84.8\,{\tiny[81.5,87.6]}} & 27.0\,{\tiny[20.7,33.7]} & 86.6\,{\tiny[81.9,90.8]} & 28.5\,{\tiny[25.2,32.0]} & 43.2\,{\tiny[39.6,46.7]} & $-$51 \\
    TG-3FM & 43.1\,{\tiny[39.3,46.4]} & \underline{95.3\,{\tiny[93.9,96.6]}} & 34.6\,{\tiny[31.2,38.3]} & \textbf{85.3\,{\tiny[82.4,88.0]}} & \underline{31.5\,{\tiny[24.5,38.9]}} & 86.2\,{\tiny[82.3,90.3]} & 26.7\,{\tiny[23.4,30.1]} & \underline{44.4\,{\tiny[40.7,48.3]}} & $-$51 \\
    TG-3FM (LoRA) & 42.5\,{\tiny[38.8,46.1]} & 93.0\,{\tiny[91.4,94.6]} & \textbf{37.4\,{\tiny[33.9,40.9]}} & 83.5\,{\tiny[80.6,86.4]} & 21.9\,{\tiny[15.7,28.4]} & \underline{88.4\,{\tiny[84.3,92.0]}} & 29.8\,{\tiny[26.5,33.3]} & 43.2\,{\tiny[39.7,46.6]} & $-$50 \\
    \midrule
    TG-H & 23.8\,{\tiny[20.5,27.1]} & 90.8\,{\tiny[88.9,92.6]} & 12.2\,{\tiny[9.9,14.4]} & 78.9\,{\tiny[75.5,82.0]} & 7.9\,{\tiny[4.5,11.8]} & 87.0\,{\tiny[82.9,90.8]} & 18.8\,{\tiny[15.9,21.9]} & 27.1\,{\tiny[24.2,30.5]} & $-$64 \\
    TG-5FM & 37.8\,{\tiny[34.2,41.6]} & 92.3\,{\tiny[90.6,93.9]} & 23.1\,{\tiny[20.3,25.9]} & 79.0\,{\tiny[76.0,82.1]} & 15.8\,{\tiny[10.8,21.1]} & \textbf{89.2\,{\tiny[85.3,92.6]}} & 28.1\,{\tiny[24.5,31.5]} & 30.9\,{\tiny[27.9,34.2]} & $-$61 \\
    TG-5FM (LoRA) & \textbf{48.5\,{\tiny[44.8,52.6]}} & 92.5\,{\tiny[90.7,94.0]} & 29.4\,{\tiny[26.0,32.5]} & 80.3\,{\tiny[77.4,83.4]} & 27.2\,{\tiny[20.4,34.0]} & 86.7\,{\tiny[81.4,91.2]} & \textbf{41.0\,{\tiny[37.0,44.8]}} & 31.9\,{\tiny[28.4,35.5]} & $-$61 \\
    \midrule
    BM25 & --- & 27.8\,{\tiny[24.2,31.4]} & --- & 23.6\,{\tiny[19.4, 27.8]} & --- & 23.0\,{\tiny[14.8, 31.2]} & --- & 32.4\,{\tiny[28.3, 36.5]} & --- \\
    te3l & --- & 47.0\,{\tiny[43.0, 51.0]} & --- & 40.6\,{\tiny[35.8, 45.4]} & --- & 38.0\,{\tiny[28.5, 47.5]} & --- & \textbf{55.6\,{\tiny[51.2, 60.0]}} & --- \\
    \bottomrule
  \end{tabular}}
  \caption{$R_c@50$ (in \%) for Gemma3-4B across all evaluation splits.
           \textbf{S1} = after Stage~1 memorization (no retrieval SFT);
           \textbf{S2} = after Stage~2 retrieval SFT (full pipeline).
           $\Delta$pp = G1$\to$RRB drop at S2.
           BM25/te3l are retrieval baselines (Hard S2 only). \textbf{Bold} results are best; \underline{underlined} are second best. Values in brackets are 95\% CIs.}
  \label{tab:recall}
\end{table*}

Table~\ref{tab:recall} reports  $R_c@50$ through both training stages and across evaluation splits. Before any retrieval SFT, Stage~1 memorization alone already reaches 38--47\% on G1 and 27--41\% on RRB for flat configurations, denoting how well the model has absorbed tool semantics during memorization. Stage~2 retrieval training then lifts G1 sharply into a 90--96\% band, consistent with~\citep{qin2025toolgen}, yet performance on RRB advances only modestly: most configurations gain 2-18pp over their Stage~1 baselines, and TG-5FM~(LoRA) actually \emph{regresses} (41.0\% $\to$ 31.9\%). The absolute G1$\to$RRB drop at Stage~2 is 50--52pp for flat and 61--64pp for hierarchical configurations, which suggests that parametric tool retrieval trained on synthetic verbose queries does not generalize to queries that differ in phrasing from the training distribution.


The ranking reverses on RRB: the best parametric model (TG-3FM, 44.4\%) falls below retrieval baseline (te3l, 55.6\%), and BM25 (32.4\%) remains competitive with several parametric configurations. Both retrieval baselines actually \emph{improve} from G1 to RRB (BM25: 27.8\%$\to$32.4\%; te3l: 47.0\%$\to$55.6\%), suggesting that the retrieval methods work better for natural-language queries.


The intermediate G2 and G3 splits trace the OOD gradient, both fall between G1 and RRB at Stage~2, confirming that the collapse is gradual rather than a sharp boundary. We examine the structural and training-dynamics reasons behind these findings in \S\ref{sec:results-is} and \S\ref{sec:results-ablations}.



\subsection{Trie-Dependency of Hierarchical Tokens}
\label{sec:results-is}

\begin{table}[t]
  \centering
  \resizebox{0.8\columnwidth}{!}{%
  \begin{tabular}{lcc}
    \toprule
    Config & G1 IS & RRB IS \\
    \midrule
    TG Gemma3-4B            & 0.75\,{\small[0.72,0.79]} & 0.85\,{\small[0.74,0.96]} \\
    TG-SP Gemma3-4B         & 0.70\,{\small[0.66,0.74]} & 0.75\,{\small[0.65,0.85]} \\
    TG-3FM Gemma3-4B        & 0.69\,{\small[0.65,0.73]} & 0.82\,{\small[0.71,0.93]} \\
    TG-3FM (LoRA) Gemma3-4B & 0.89\,{\small[0.86,0.93]} & 0.81\,{\small[0.71,0.91]} \\
    \midrule
    TG-H Gemma3-4B          & 0.42\,{\small[0.38,0.46]} & 0.33\,{\small[0.24,0.41]} \\
    TG-5FM Gemma3-4B        & 0.84\,{\small[0.80,0.87]} & 0.64\,{\small[0.53,0.76]} \\
    TG-5FM (LoRA) Gemma3-4B & 0.66\,{\small[0.62,0.70]} & 0.79\,{\small[0.65,0.92]} \\
    \midrule
    TG-SP Qwen3.5-4B        & 0.90\,{\small[0.86,0.93]} & 0.98\,{\small[0.89,1.00]} \\
    TG-SP (LoRA) Qwen3.5-4B & \textbf{0.94\,{\small[0.92,0.97]}} & \textbf{0.99\,{\small[0.89,1.00]}} \\
    TG-3FM Qwen3.5-4B       & 0.79\,{\small[0.75,0.83]} & 0.81\,{\small[0.73,0.89]} \\
    TG-3FM (LoRA) Qwen3.5-4B & 0.68\,{\small[0.64,0.72]} & 0.82\,{\small[0.73,0.92]} \\
    TG-H Qwen3.5-4B         & 0.57\,{\small[0.53,0.61]} & 0.28\,{\small[0.21,0.35]} \\
    TG-SP Gemma3-12B        & 0.66\,{\small[0.62,0.70]} & 0.84\,{\small[0.75,0.93]} \\
    TG-3FM (LoRA) Gemma3-12B & 0.76\,{\small[0.72,0.80]} & 0.86\,{\small[0.76,0.96]} \\
    \bottomrule
  \end{tabular}}
  \caption{Internalization Score (IS@50) after Stage~2 training. 95\% CI in brackets.}
  \label{tab:is}
\end{table}

\begin{figure}[t]
  \centering
  \includegraphics[width=\linewidth]{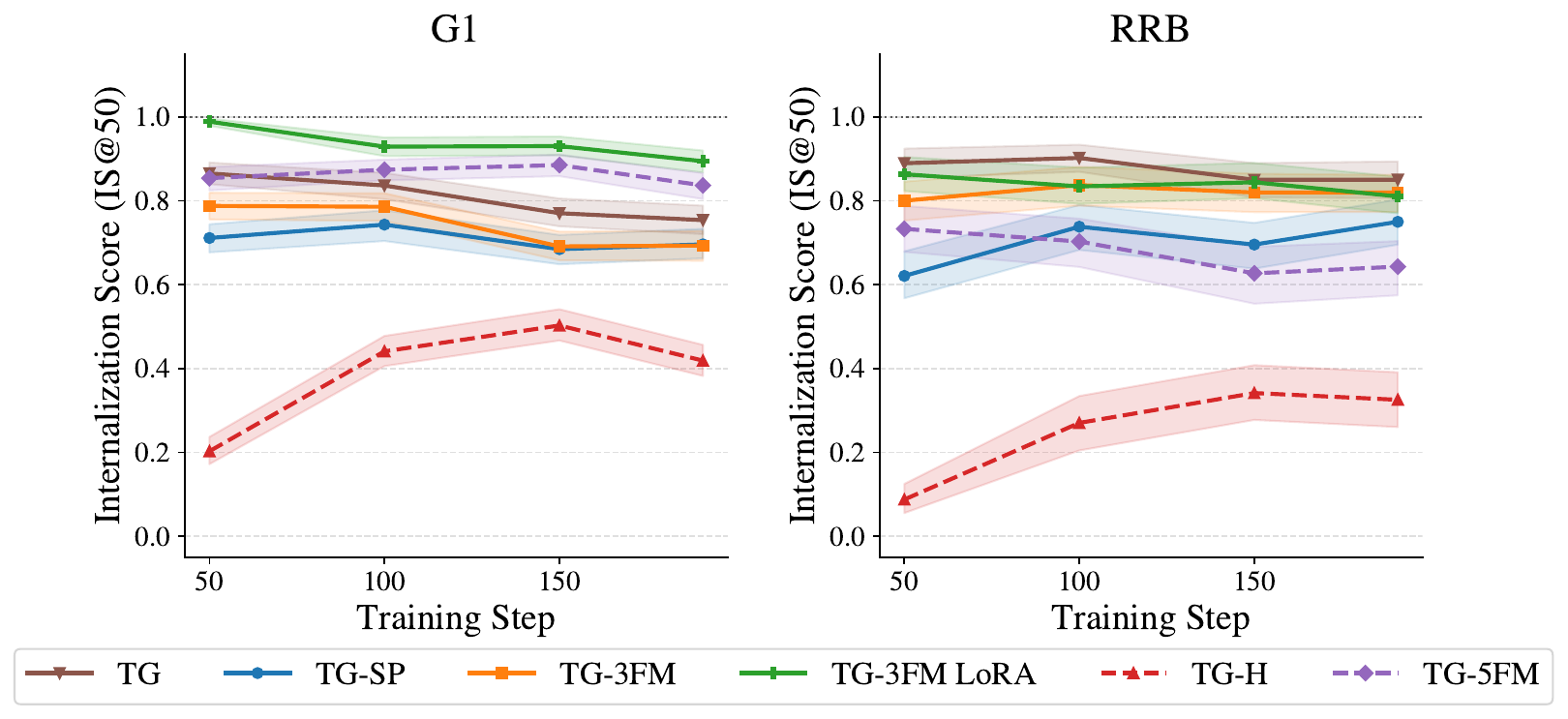}
  \caption{IS@50 across Stage~2 training steps for Gemma3-4B. Shaded bands are 95\% bootstrap CIs. Full results across all splits \& models are in Appendix~\ref{sec:appendix-is-full}.}
  \label{fig:is-evolution}
\end{figure}

The Internalization Score quantifies whether a model generates valid tool tokens independently or relies on trie guidance for heavy lifting (IS$\approx$1.0: independent; IS$\approx$0: trie-dependent).

Table~\ref{tab:is} reports IS after Stage~2 ($B=50$). The separation between token formats is large: flat configurations achieve RRB IS in the range 0.75--0.85, while hierarchical configurations fall to 0.33--0.79, a gap of Cohen's $d = 1.45$~\citep{Cohen1988}, indicating the two groups occupy fundamentally different regions of trie-dependency. The gap is \emph{larger} on RRB than on G1 for hierarchical tokens (TG-H: G1=0.42, RRB=0.33), while flat tokens maintain or improve IS on RRB (TG: G1=0.75, RRB=0.85), suggesting that the distributional shift compounds with structural trie-dependency. The flat-over-hierarchical IS gap holds across architectures: Qwen3.5-4B flat configurations reach IS$\geq$0.98 on RRB, while TG-H Qwen3.5-4B drops to 0.28; full IS results across all architectures are in Appendix~\ref{sec:appendix-is-full}.


Figure~\ref{fig:is-evolution} traces IS evolution: TG-H stays below IS$=$0.35 throughout Stage2 training, while TG-3FM(LoRA) declines only modestly (1.0$\to$0.89), confirming LoRA preserves free-generation capacity from Stage~1.

\subsection{Knowledge-Retrieval Dissociation}
\label{sec:results-probes}

Table~\ref{tab:probes} reports MCQ and QA probing accuracy corresponding to Gemma3-4B model after Stage~1 and Stage~2 trainings. The headline finding is a striking dissociation: a model (TG) achieving $\sim$95\% $R_c@50$ on G1 scores only 31.4\% on 4-way MCQ (random=25\%) and exactly 50.0\% on binary QA (random=50\%), suggesting the behavior to be more like a lookup table, not a knowledge base. At Stage~2, full-finetuned models score 20--31\% on MCQ and 34--50\% on QA; LoRA extends the upper end to 41.7\% MCQ and 56.4\% QA.

The picture differs sharply by token format. Flat configurations accumulate genuine knowledge during Stage~1 (TG enters retrieval training at 55.4\% MCQ and TG-3FM at 32.9\%), which Stage~2 then systematically erodes (TG: $-$24.0pp; TG-3FM: $-$3.1pp). Hierarchical tokens tell a different story: TG-H and TG-5FM score \emph{below} random chance at Stage~1 itself (23.6\% [19.8, 27.2] and 22.8\% [18.9, 26.6]), before any retrieval training has occurred, suggesting that these configurations never established semantic grounding in the first place. LoRA substantially buffers Stage~2 destruction for configurations where knowledge does take hold: TG-3FM~(LoRA) retains 41.7\% MCQ (CI [37.7, 46.2]) vs.\ FFT at 29.8\% (CI [25.6, 33.9]) and achieves the highest QA score at Stage~2 (56.4\% [52.2, 60.6]).

\begin{figure}[t]
  \centering
  \includegraphics[width=0.80\linewidth]{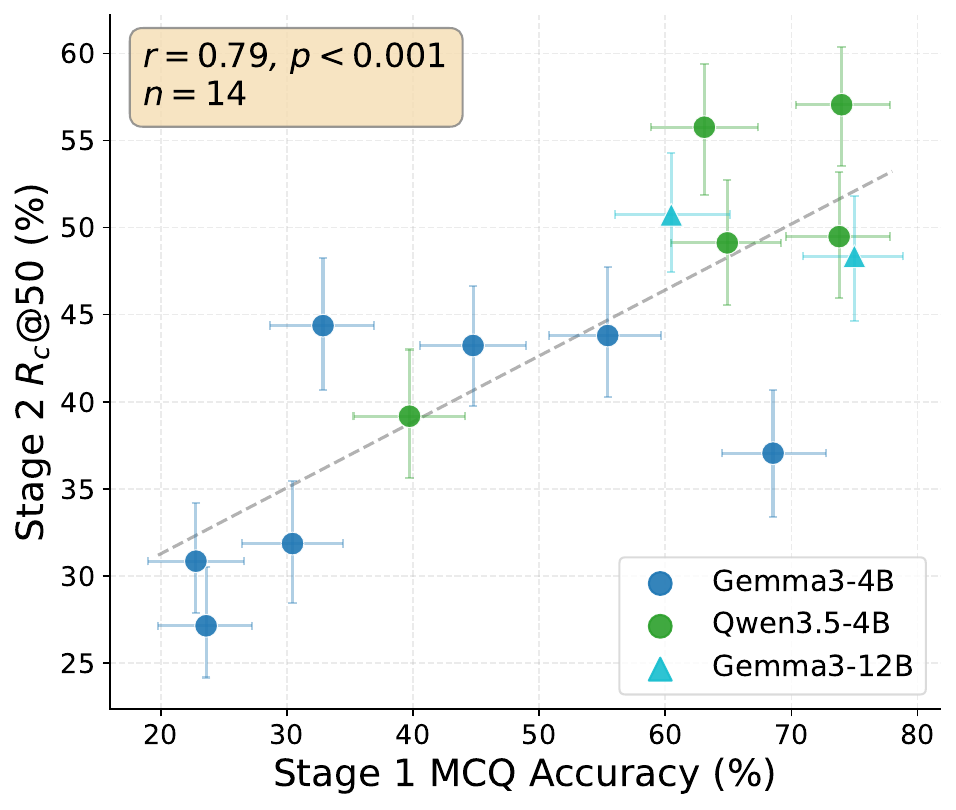}
  \caption{Correlation between Stage~1 MCQ accuracy and Stage~2 RRB $R_c@50$ across all model families ($n=14$, $r=0.79$, $p<0.001$). Error bars show 95\% CIs. All 14 model variants are listed in Table~\ref{tab:correlation-variants} (Appendix~\ref{sec:appendix-probes-scale}).}
  \label{fig:correlation}
\end{figure}

Across all configurations and model families, Stage~1 MCQ accuracy strongly correlates to Stage~2 RRB $R_c@50$ (Pearson $r = 0.79$, $p < 0.001$, $n = 14$; Figure~\ref{fig:correlation}): configurations that enter retrieval training with richer semantic representations generalize better to OOD queries, even after partial knowledge destruction. (cross-architecture results in Appendix~\ref{sec:appendix-probes-scale}).

This correlation also validates the probes as a principled proxy for downstream capability: rather than measuring end-to-end task execution directly using parametric tool selection, which itself is an open training problem we leave to future work (\S\ref{sec:limitations}), we use MCQ/QA to quantify the semantic grounding on which downstream agentic fine-tuning (planning, multi-step orchestration, response generation over private APIs) depends~\citep{qin2025toolgen}. A model whose grounding collapses to near-random after Stage~2 has lost the very property that would make it a useful foundation for those tasks, whereas Stage~1 MCQ predicts which configurations retain it.


\begin{table}[t]
  \centering
  \resizebox{\columnwidth}{!}{%
  \begin{tabular}{lcccc}
    \toprule
    Config & \multicolumn{2}{c}{MCQ (S1$\rightarrow$S2)} & \multicolumn{2}{c}{QA (S1$\rightarrow$S2)} \\
    \cmidrule(lr){2-3} \cmidrule(lr){4-5}
           & S1 & S2 & S1 & S2 \\
    \midrule
    TG            & \textbf{55.4\,{\small[50.8,59.7]}} & 31.4\,{\small[27.6,35.5]} & 57.8\,{\small[53.6,62.2]} & 50.0\,{\small[45.8,54.0]} \\
    TG-3FM        & 32.9\,{\small[28.6,36.9]} & 29.8\,{\small[25.6,33.9]} & 60.2\,{\small[55.8,64.6]} & 50.2\,{\small[45.8,54.8]} \\
    TG-3FM (LoRA) & 44.8\,{\small[40.5,49.0]} & \textbf{41.7}\,{\small[37.7,46.2]} & \textbf{79.8\,{\small[76.0,83.2]}} & \textbf{56.4\,{\small[52.2,60.6]}} \\
    TG-H          & 23.6\,{\small[19.8,27.2]} & 20.4\,{\small[16.9,23.8]} & 50.6\,{\small[46.2,55.2]} & 34.4\,{\small[30.0,39.0]} \\
    TG-5FM        & 22.8\,{\small[18.9,26.6]} & 21.6\,{\small[17.7,25.2]} & 50.2\,{\small[46.0,54.8]} & 48.4\,{\small[44.2,52.6]} \\
    TG-5FM (LoRA) & 30.4\,{\small[26.4,34.5]} & 28.8\,{\small[25.0,32.9]} & 56.2\,{\small[52.0,60.6]} & 52.2\,{\small[47.8,56.4]} \\
    \bottomrule
  \end{tabular}}
  \caption{MCQ and QA probing accuracy (Gemma3-4B). Random: MCQ=25\%, QA=50\%. 95\% CI in brackets.}
  \label{tab:probes}
\end{table}

\subsection{Design Choice Ablations}
\label{sec:results-ablations}

We isolate which design choices mitigate or exacerbate the three findings above.

\textbf{Token format.} Flat tokens (TG-SP, Gemma3-4B) achieve RRB $R_c@50$=43.2\% (95\% CI [39.6, 46.7]) and IS=0.75, vs.\ hierarchical (TG-H, Gemma3-4B) at RRB $R_c@50$=27.1\% (CI [24.2, 30.5]) and IS=0.33 (Cohen's $d = 8.23$ on recall, $d = 1.45$ on IS~\citep{Cohen1988}). This replicates across architectures: Qwen3.5-4B flat $R_c@50$=55.8\% vs.\ hier $R_c@50$=39.2\% (+16.6pp). Flat tokens require a single decoding step; hierarchical tokens require multi-step composition where errors at the API level propagate to the endpoint level.

\begin{figure}[t]
  \centering
  \includegraphics[width=\linewidth]{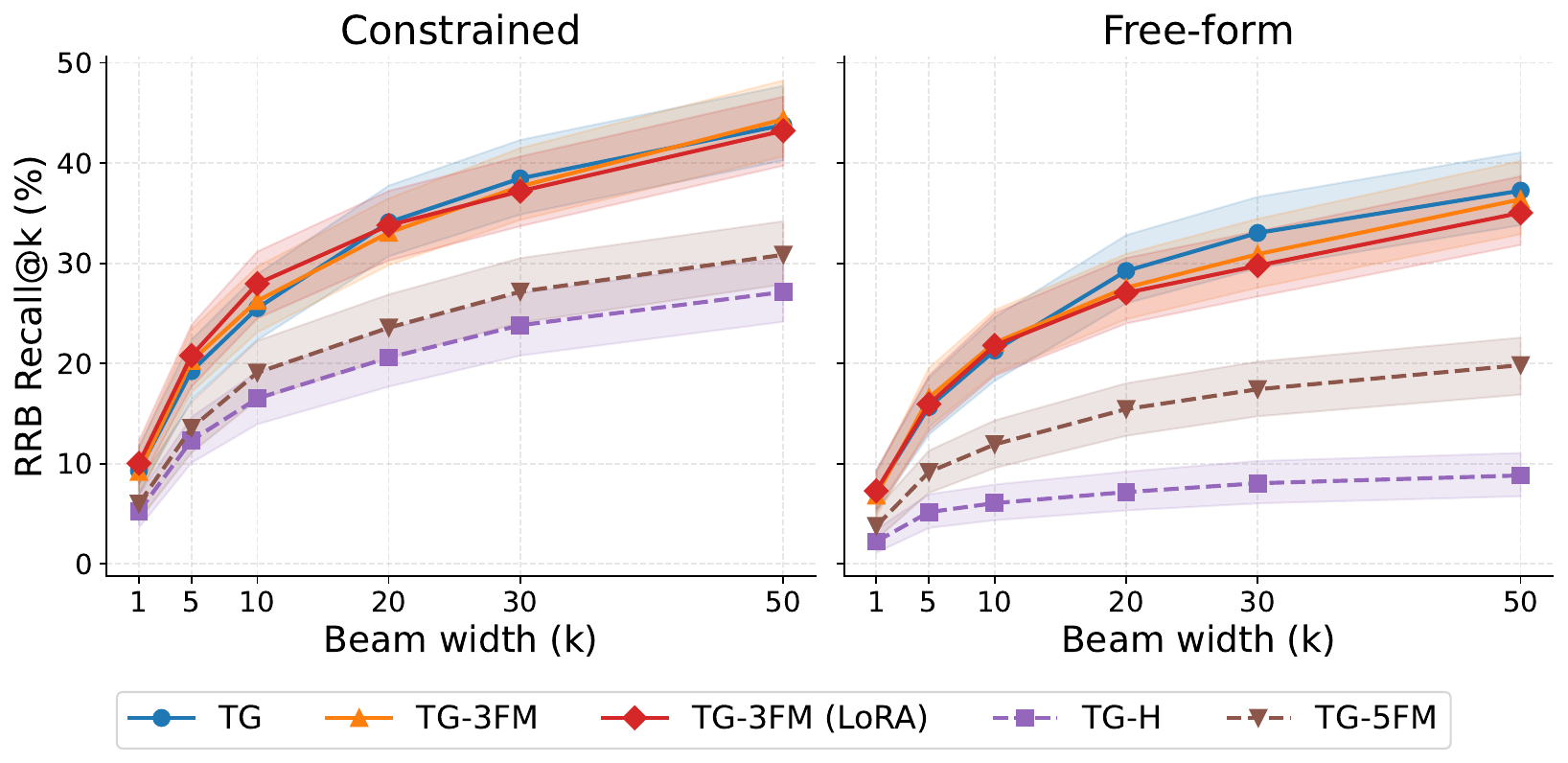}
  \caption{RRB $R_c@k$ and $R_f@k$ vs.\ beam width $k$ for five Gemma3-4B Stage~2 configurations. Shaded bands are 95\% CIs.}
  \label{fig:recall-scaling}
\end{figure}

Figure~\ref{fig:recall-scaling} shows the gap persists across all beam widths: TG-H plateaus below 9\% $R_f@k$ regardless of beam budget, while flat configs scale to 37\%.

\begin{figure}[t]
  \centering
  \includegraphics[width=0.90\linewidth]{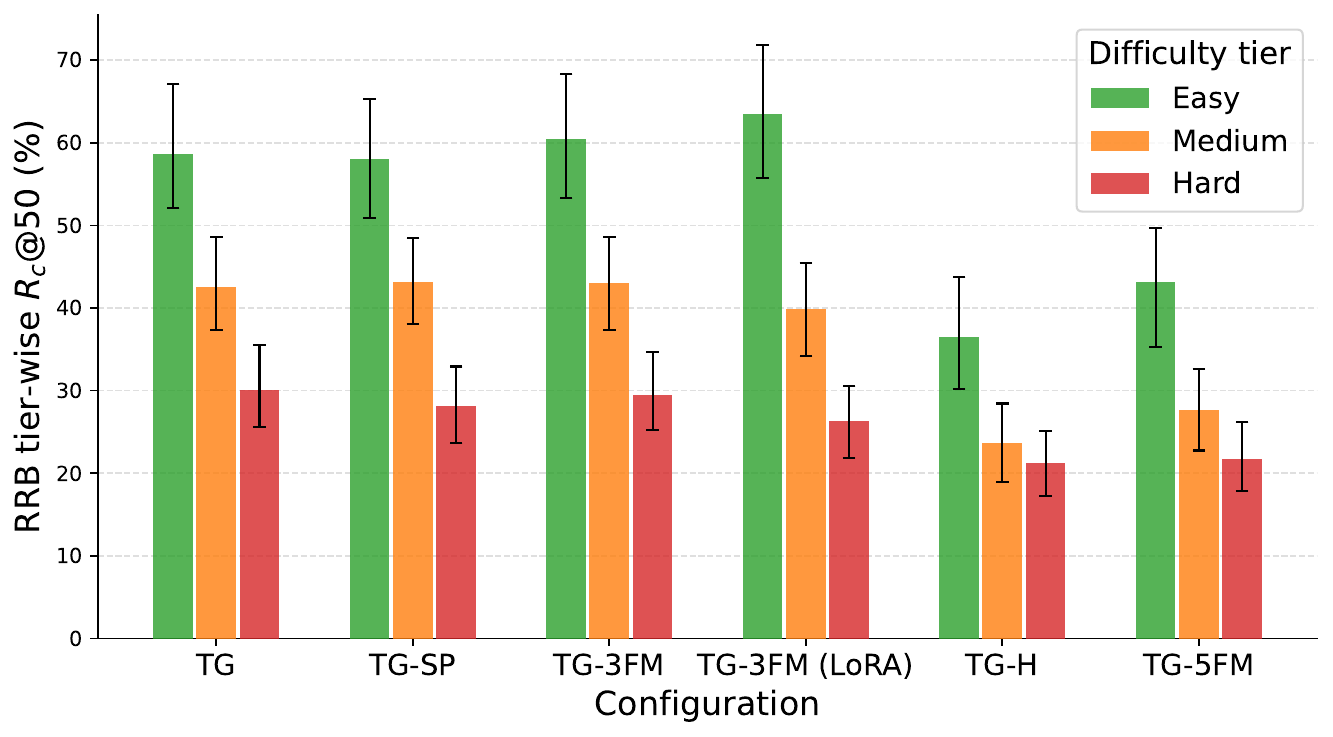}
  \caption{$R_c@50$ by RRB difficulty tier for five Gemma3-4B Stage~2 configurations. Error bars are 95\% bootstrap CIs.}
  \label{fig:rrb-tiers}
\end{figure}

Figure~\ref{fig:rrb-tiers} shows flat configurations declining progressively across RRB difficulty tiers (Easy 58--64\% $\to$ RRB 26--30\%), while hierarchical configurations underperform at every level (Easy: 37--43\%, RRB: 21--22\%). Token format is the dominant separator, flat leads by 15--25~pp on Easy and Medium, with TG-3FM~(LoRA) achieving the highest single-target retrievability (Easy: 63.5\%).

\textbf{LoRA as knowledge preservation.} For Gemma3-4B TG-3FM, LoRA preserves MCQ at 41.7\% (95\% CI [37.7, 46.2]) vs.\ FFT at 29.8\% (CI [25.6, 33.9]). For TG-5FM, the trend is 28.8\% (CI [25.0, 32.9]) vs.\ 21.6\% (CI [17.7, 25.2]). For hierarchical tokens, LoRA also improves IS: TG-5FM~(LoRA) achieves RRB IS=0.79 vs.\ FFT at 0.64. The effect is most pronounced where FFT causes severe destruction; for Qwen3.5-4B, which already preserves knowledge without LoRA (TG-3FM: 72.8\%), adding LoRA yields only marginal additional gain (74.2\%).

\textbf{Virtual tool token embedding drift.}
We measure relative L2 drift of virtual tool tokens vs.\ base-vocabulary tokens across Stage~1$\to$2: virtual tool tokens drift $2$--$3\times$ farther under FFT and $22.9\times$ farther under LoRA with embedding layer fully-finetuned (Mann-Whitney $p < 2{\times}10^{-308}$), yet within-cluster cosine similarity barely changes ($|\Delta\mathrm{cosim}| < 0.002$), suggesting tokens shift as coherent clusters rather than diverging to encode distinct knowledge content.
Hierarchical tokens enter Stage~2 near-orthogonal to each other ($\mathrm{cosim}_{S1}{\approx}0.038$ vs.\ $0.37$ for flat tokens) and require $7$--$24\times$ less repositioning, yet still fail MCQ probes just as severely — suggesting that token geometry is not the source of the knowledge-retrieval dissociation (full analysis in Appendix~\ref{sec:appendix-emb-drift})

\begin{figure}[t]
  \centering
  \includegraphics[width=\linewidth]{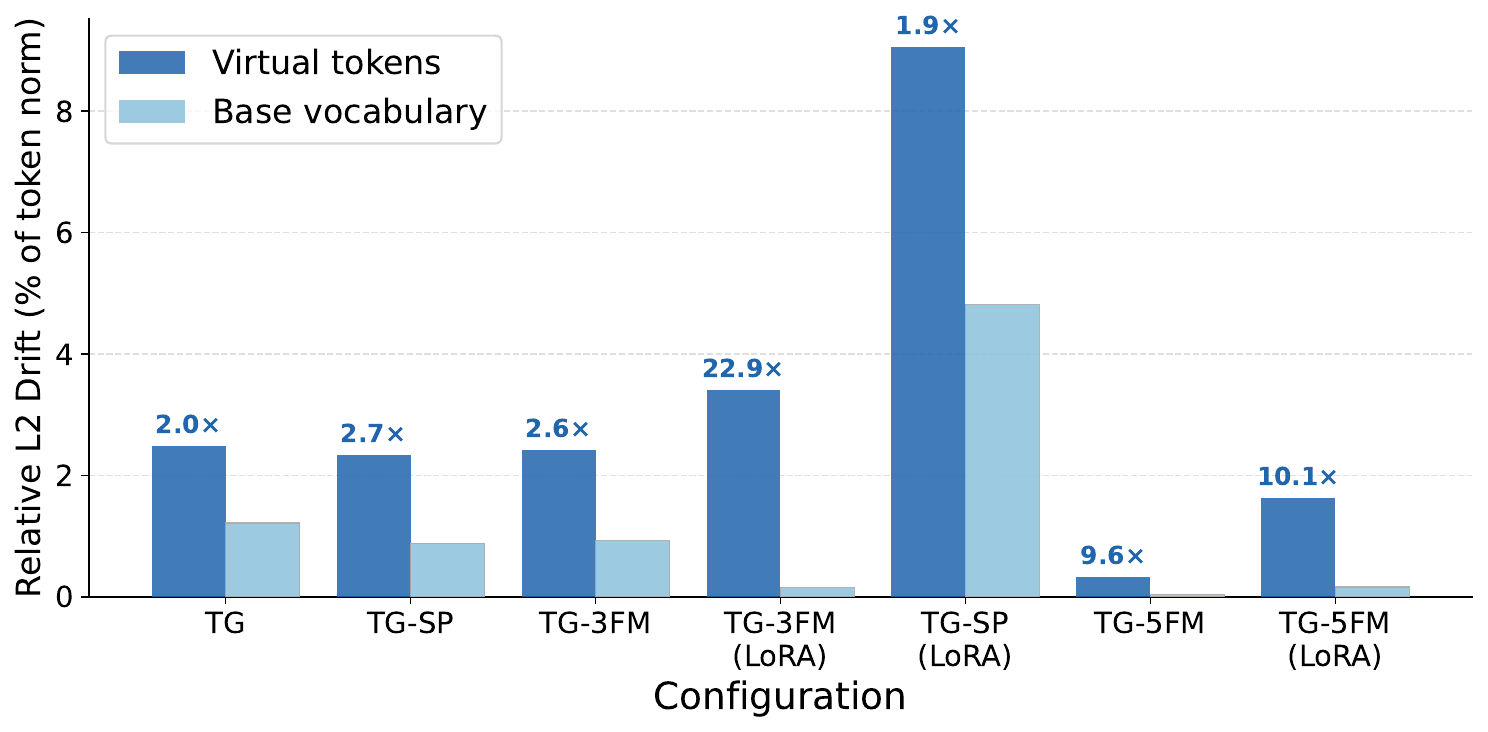}
  \caption{Relative L2 drift ($d_\mathrm{rel}$) of virtual vs.\ randomly sampled vocabulary tokens across Stage~1$\to$Stage~2 for Gemma3-4B configurations.}
  \label{fig:emb-drift}
\end{figure}

\textbf{Multi-format training.} Multi-format memorization TG-3FM combined with LoRA achieves the highest Stage~2 MCQ accuracy across all architectures: 41.7\% (Gemma3-4B), 74.2\% (Qwen3.5-4B), 76.4\% (Gemma3-12B). Seeing the same tool tokens in multiple different contexts during Stage~1 appears to build relatively more robust representations that are harder to overwrite during Stage~2 retrieval fine-tuning.

\textbf{System prompt.} During Stage~2, TG (no SP) achieves RRB $R_c@50$=43.8\% (95\% CI [40.3, 47.7]) vs.\ TG-SP at 43.2\% (CI [39.6, 46.7]), fully overlapping intervals, suggesting that the system prompt has no measurable effect once retrieval training is complete.


\textbf{Training data distribution.}
To test whether the knowledge-retrieval dissociation is a consequence of Stage~2 query distribution rather than the training objective itself, we trained a TG variant whose Stage~2 data consists entirely of RRB-style queries generated by ToolSense RRB generator over the same $\sim$47k tool catalog.
Retraining TG Stage~2 on RRB-style queries (disjoint from evaluation set; see Appendix~\ref{sec:appendix-realistic-training}) shows that RRB $R_c@50$ rises from 43.8\% to 87.8\% ($+$44.0pp), but G1 $R_c@50$ drops from 95.7\% to 86.6\%, suggesting that the model over-specializes to query style that appears in Stage~2 training.
MCQ accuracy drops from 31.4\% to 26.0\%, almost near-random for a 4-choice task.
The knowledge-retrieval dissociation thus persists regardless of Stage~2 training data distribution, pointing to the SFT objective itself as the deeper root cause.

\subsection{Cross-Architecture Validation}
\label{sec:results-cross-arch}



Table~\ref{tab:cross-arch} confirms both findings generalize: the RRB-recall collapse persists across all model families, and flat tokens consistently outperform hierarchical by 16--18~pp (IS: 0.75--0.98 flat vs.\ 0.28--0.33 hierarchical). Knowledge retention is not explained by model capacity alone: Qwen3.5-4B retains 62--73\% MCQ despite matching Gemma3-4B in size (29--31\%), and Gemma3-12B TG-3FM LoRA variant achieves the highest MCQ retention (76.4\%) with competitive RRB recall (48.4\%).

\begin{table}[t]
  \centering
  \tiny
  \begin{tabular}{@{}llccc@{}}
    \toprule
    Model & Config & RRB Recall & IS & MCQ \\
    \midrule
    \multirow{4}{*}{Gemma3-4B}
      & TG-SP         & 43.2 & 0.75 & 31.4 \\
      & TG-3FM        & 44.4 & 0.82 & 29.8 \\
      & TG-3FM (LoRA) & 43.2 & 0.81 & 41.7 \\
      & TG-H          & 27.1 & 0.33 & 20.4 \\
    \midrule
    \multirow{4}{*}{Qwen3.5-4B}
      & TG-SP         & 55.8 & \textbf{0.98} & 61.7 \\
      & TG-3FM        & \textbf{57.0} & 0.81 & 72.8 \\
      & TG-3FM (LoRA) & 49.5 & 0.82 & 74.2 \\
      & TG-H          & 39.2 & 0.28 & 31.6 \\
    \midrule
    \multirow{2}{*}{Gemma3-12B}
      & TG-SP         & 50.7 & 0.84 & 66.3 \\
      & TG-3FM (LoRA) & 48.4 & 0.86 & \textbf{76.4} \\
    \bottomrule
  \end{tabular}
  \caption{Cross-architecture results (Stage~2, beam=50).}
  \label{tab:cross-arch}
\end{table}

\subsection{Interpreting ToolSense Scores}
\label{sec:results-interpreting}

Having presented the full results, we distill the diagnostics into actionable reference points, calibrated against the configurations studied above. For factual probing, Stage~1 MCQ accuracy $>40\%$ indicates that a configuration has established genuine semantic grounding (e.g., TG at 55.4\% and TG-3FM~(LoRA) at 44.8\%; Table~\ref{tab:probes}), whereas accuracy near the 25\% random baseline signals that no grounding was acquired. For trie-dependence, IS$>0.65$ indicates sufficient trie-independence to support free-form retrieval, while IS$<0.35$ indicates collapse, the tool token can only be produced under trie guidance (e.g., TG-H on RRB; Table~\ref{tab:is}). These thresholds are diagnostic guides, not hard cutoffs, and should be read alongside the reported confidence intervals.

\section{Conclusion}
\label{sec:conclusion}

We introduced \textbf{ToolSense}, a diagnostic framework that auto-generates three benchmarks---RRB, MCQ, and QA---from any tool catalog to probe parametric retrieval systems along orthogonal axes: OOD generalization, trie-dependency, and factual knowledge retention.
Applying ToolSense to ToolBench reveals that high constrained recall on verbose in-distribution queries is a poor proxy for real-world capability: performance collapses on realistic queries, hierarchical tokens exhibit deep trie-dependency, and Stage~2 retrieval fine-tuning systematically erodes the tool knowledge built during Stage~1, leaving models that behave more like lookup tables than knowledge bases.
LoRA combined with multi-format Stage~1 memorization best mitigates these failure modes, and Stage~1 MCQ accuracy is a reliable early predictor of downstream OOD generalization.
We release the ToolSense framework and the ToolBench diagnostic benchmarks to encourage further research.

\section*{Limitations}
\label{sec:limitations}

While ToolSense's benchmarks are designed to generalize to other parametric retrieval designs, our empirical study focuses on ToolGen's two-stage paradigm~\citep{qin2025toolgen} applied to ToolBench; extending the diagnosis to other catalog-agnostic parametric designs is a natural next step.
The RRB, MCQ, and QA benchmarks are LLM-generated and validated by human annotators on 100 samples each ($\kappa \geq 0.805$); scaling the human study beyond 100 samples per benchmark would further strengthen confidence in the benchmark quality, though doing so at catalog scale involves non-trivial annotation cost and time.
Our analysis focuses on the retrieval stage in isolation; end-to-end agentic evaluation, where retrieval quality interacts with planning and execution, is a promising direction for future work.
The open-source base models we study (Gemma3, Qwen) may have encountered RapidAPI tool documentation during pre-training, potentially inflating absolute Stage~1 knowledge scores relative to truly novel tools; the relative forgetting patterns across Stage~1$\to$Stage~2 remain unaffected by this confound, though our findings may not directly transfer to private enterprise tool catalogs whose APIs never appear in any public pre-training corpus.
The Internalization Score is a ratio metric and can exhibit high variance when $R_c@k$ is low; while we discuss and handle the degenerate $R_c@k=0$ case explicitly and reported confidence intervals already capture this variance, future work should evaluate log-ratio or calibration-based formulations to confirm the stability of IS findings across catalog sizes.
Finally, our model experiments cover the 4B--12B parameter range; studying very large models ($\geq$30B) may reveal additional nuances in how model scale affects knowledge retention under sequential fine-tuning.

\section*{Ethical Considerations}
\label{sec:ethics}

We conducted experiments within the provisions of the ACL Ethics Policy and relevant research-integrity guidelines. There are, to the best of our knowledge, no remaining ethical risks that have not been addressed.

\bibliography{custom}

\appendix

\section{ToolBench Standard Splits vs.\ RRB: Query Style Comparison}
\label{sec:appendix-query-comparison}

\begin{figure*}[h]
  \centering
  \includegraphics[width=\linewidth]{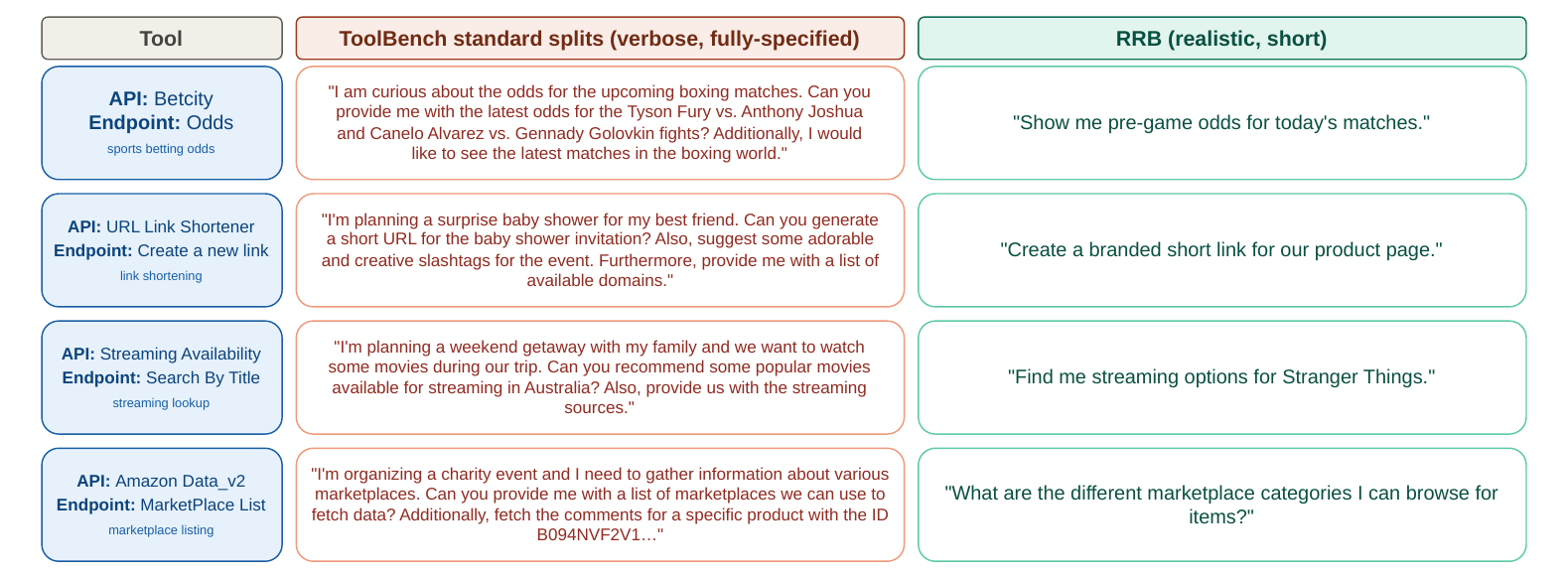}
  \caption{Side-by-side comparison of ToolBench standard evaluation split queries (G1/G2/G3) and
    RRB queries for the same tools. ToolBench queries are verbose, story-wrapped, and
    enumerate specific parameters. Whereas, RRB queries are short, intent-focused, and reflect how a real user
    would naturally phrase a request.}
  \label{fig:query-comparison}
\end{figure*}

Figure~\ref{fig:query-comparison} illustrates a fundamental difference in query style
between the ToolBench standard evaluation splits (G1/G2/G3) and our Realistic Retrieval
Benchmark (RRB). For the same underlying tools, ToolBench queries are verbose
narrative prompts that effectively paraphrase the API documentation.
A model that has memorized tool descriptions at Stage~1 can retrieve the correct
token by matching surface-level vocabulary, without genuine semantic understanding.

RRB queries, by contrast, are generated by \(\theta_{\text{RRB}}\) with an
explicit instruction to produce \emph{intent-focused} natural language: a concise
statement of what the user wants to accomplish, stripped of API-specific terminology.
The four examples shown span distinct domains, yet in each case, the RRB query is how a real user
would type the request into a chat interface: ``Show me pre-game odds for today's
matches,'' ``Create a branded short link for our product page,'' ``Find me streaming
options for Stranger Things,'' and ``What are the different marketplace categories I can
browse for items?''
This shift from documentation-paraphrase to user-intent query is precisely the
distribution gap that causes retrieval collapse in our experiments
(Section~\ref{sec:results}): a parametric model trained exclusively on
Stage~2 (query $\to$ virtual token) pairs derived from verbose ToolBench-style prompts
has not learned to bridge the lexical gap to short, colloquial, intent-focused user utterances.

\section{Multi-Choice Tool Selection (MCTS) Memorization Dataset}
\label{sec:appendix-mcts}

The standard Stage~1 memorization format trains the model on
(tool metadata $\to$ virtual token) pairs~\citep{qin2025toolgen}.
While effective at associating a tool with its token, this objective does not explicitly
teach the model to \emph{distinguish} semantically similar tools --- a critical
requirement in large catalogs where many tools share overlapping descriptions.

\paragraph{Format.}
MCTS is a discriminative Stage~1 objective.
For each tool $\tau$, we construct a multiple-choice instance by pairing its description
$\mathrm{desc}(\tau)$ with $K$ hard-negative tool descriptions
$\mathcal{H}_K(\tau) = \{\tau_1^-, \ldots, \tau_K^-\}$ retrieved by
embedding-based similarity.
At training time, the model receives the description of $\tau$ alongside $K{+}1$ tool tokens corresponding to the candidates (shuffled) and must generate the virtual token $v_\tau$ corresponding to the ground-truth tool.
The training signal is therefore a next-token loss conditioned on a \emph{contrast set}, forcing the model to resolve fine-grained semantic boundaries rather than simply recall a memorized association.

\paragraph{Hard-negative mining.}
Candidates are retrieved using a dense embedding model (\texttt{text-embedding-3-large})
indexed over the full tool catalog via chromadb.
For each anchor tool, we over-fetch $K{+}5$ nearest neighbors by cosine similarity
and remove the anchor itself, yielding a pool of $K$ hard negatives.
Because these tools are semantically proximate but functionally distinct
(e.g., two overlapping enterprise search APIs, or two QR-code generation services),
they represent the most plausible confusables --- exactly the cases where
a model relying on shallow lexical overlap would fail. For our experiments, we set $K=4$.

\paragraph{Effect on Stage~1 representations.}
By requiring the model to correctly identify $v_\tau$ against its nearest semantic
neighbors, MCTS encourages the embedding of $v_\tau$ to be positioned not just
\emph{near} its own description, but \emph{farther} from confusable alternatives.
This discriminative geometry is more resistant to the overwriting that occurs during
Stage~2 retrieval SFT, as reflected in the higher Stage~2 MCQ scores observed for
configurations that include MCTS in their Stage~1 training mix
(TG-3FM, TG-5FM; Table~\ref{tab:configs}).

\section{Training Configuration Details}
\label{sec:appendix-configs}

\subsection{Stage~1: Memorization Formats}
\label{sec:appendix-s1-formats}

Stage~1 fine-tunes the model to associate each tool's virtual token with its metadata.
We define five memorization formats, each presenting the task from a different angle.
All formats share the same tool catalog but differ in what is given as input and what
the model is trained to generate. For applicable model configurations, exact system prompts are provided in Appendix~\ref{sec:appendix-prompts-memo}.

\paragraph{desc$\to$tok (standard).}
The model receives the full tool description and is trained to generate the virtual token.
This is the format introduced by \citet{qin2025toolgen} and used in \emph{all} configurations,
regardless of token scheme.
The form of \texttt{tok} differs by scheme: for flat tokens it is a single atomic identifier
encoding the full tool identity (\texttt{<<API\&\&ENDPOINT>>}); for hierarchical tokens it is
the joint two-token sequence (\texttt{<<API>><<ENDPOINT>>}), generated autoregressively as
two successive decode steps.
\begin{quote}\small
  \textit{Input:} ``Manages and administers search queries within the Enterprise Search
  Administration system\ldots'' \\
  \textit{Target (flat):} \texttt{<<ESH\_ADMIN\_SRV\_0001\&\&SearchQueries>>} \\
  \textit{Target (hierarchical):} \texttt{<<ESH\_ADMIN\_SRV\_0001>><<SearchQueries>>}
\end{quote}

\paragraph{tok$\to$desc (reverse).}
The input and target are swapped: the model receives the virtual token and is trained to
generate the tool description.
This forces bidirectional association---the model must encode not only
\emph{which description maps to which token} but also \emph{which token maps to which
description}---making the embedding more resistant to overwriting during Stage~2.
\begin{quote}\small
  \textit{Input (flat):} \texttt{<<ESH\_ADMIN\_SRV\_0001\&\&SearchQueries>>} \\
  \textit{Input (hierarchical):} \texttt{<<ESH\_ADMIN\_SRV\_0001>><<SearchQueries>>} \\
  \textit{Target:} ``Manages and administers search queries\ldots''
\end{quote}

\paragraph{desc$\to$api\_tok (hierarchical only).}
The model receives the tool description and is trained to generate only the API-level token---the first step of the two-token hierarchical sequence.
This format is specific to hierarchical token configuration TG-5FM and provides an
additional supervised signal that trains the first generation step in isolation,
supplementing the primary \texttt{desc$\to$tok} format which already trains both tokens
jointly end-to-end.
\begin{quote}\small
  \textit{Input:} ``Manages and administers search queries\ldots'' \\
  \textit{Target:} \texttt{<<ESH\_ADMIN\_SRV\_0001>>}
\end{quote}

\paragraph{desc+api\_tok$\to$endpoint\_tok (hierarchical only).}
The model receives the tool description with the API-level token and is
trained to generate the endpoint-level token, the second step of the hierarchical
sequence.
Together with desc$\to$api\_tok, this provides two additional supervised signals that
train each hierarchical step independently, complementing the joint
\texttt{desc$\to$tok} objective already present in TG-5FM.
This decomposition reinforces each level of the hierarchy with a dedicated loss signal.
\begin{quote}\small
  \textit{Input:} ``Manages and administers search queries\ldots{}
  \texttt{<<ESH\_ADMIN\_SRV\_0001>>}'' \\
  \textit{Target:} \texttt{<<SearchQueries>>}
\end{quote}

\paragraph{MCTS (discriminative).}
See Appendix~\ref{sec:appendix-mcts} for a full description.
In brief, the model must identify the correct virtual token from a set of $K+1$ tool tokens, which include ground truth tool token as well as $K$ hard-negative candidates retrieved by embedding similarity, adding a discriminative objective alongside the generative formats above.

\subsection{Stage~2: Retrieval Training Data}
\label{sec:appendix-s2-data}

Stage~2 fine-tunes the model on (query $\to$ virtual token) pairs drawn from the
ToolBench training split~\citep{qin2023toolllm}, following \citet{qin2025toolgen}.
The Stage~2 data is \emph{identical} across all five configurations; no format-specific
data augmentation is applied at this stage.
The only variation is whether a system prompt is prepended to each training instance:
TG omits it entirely, while all other configurations (TG-SP, TG-3FM, TG-H, TG-5FM)
include the same task-specific system prompt used during Stage~1, ensuring
consistency between the training stages and inference. For applicable model configurations, exact system prompts are provided in Appendix~\ref{sec:appendix-prompts-retrieval}.

\subsection{Design Rationale per Configuration}
\label{sec:appendix-config-rationale}

The five configurations isolate one variable at a time against a shared baseline:

\begin{itemize}[topsep=2pt,itemsep=2pt]
  \item \textbf{TG vs.\ TG-SP}: isolates the effect of the system prompt, all else equal.
  \item \textbf{TG-SP vs.\ TG-3FM}: isolates the effect of enriching Stage~1 with
        reverse (tok$\to$desc) and discriminative (MCTS) objectives, holding token
        format and system prompt constant.
  \item \textbf{TG-SP vs.\ TG-H}: isolates the effect of switching from flat to
        hierarchical tokens, holding memorization format (desc$\to$tok only) and
        system prompt constant.
  \item \textbf{TG-H vs.\ TG-5FM}: isolates the effect of adding all five
        memorization formats to the hierarchical token baseline.
  \item \textbf{FFT vs.\ LoRA} (within any configuration): isolates the effect of
        constraining parameter update magnitude on knowledge retention across
        the sequential training stages~\citep{biderman2024lora}.
\end{itemize}


\section{Training Hyperparameters}
\label{sec:appendix-hyperparams}

All experiments were run on a single node with 8 NVIDIA H200 GPUs. Each model was trained on a single GPU, with the AdamW optimizer, bf16 mixed precision, and a maximum sequence length of 1024 tokens.
The effective batch size is 1024 for all runs (achieved via gradient accumulation: 1 GPU $\times$ 2 samples $\times$ 512 accumulation steps for 4B models; 2 GPUs $\times$ 2 $\times$ 256 for 12B models).

\begin{table}[h]
\centering
\small
\begin{tabular}{lcc}
\toprule
Hyperparameter & FFT & LoRA \\
\midrule
\multicolumn{3}{l}{\textit{Stage 1 — Memorization (2 epochs)}} \\
\quad Learning rate    & $2{\times}10^{-5}$ & $1{\times}10^{-4}$ \\
\quad LR scheduler     & Cosine & Cosine \\
\quad Min LR floor     & $2{\times}10^{-6}$ & $1{\times}10^{-5}$ \\
\midrule
\multicolumn{3}{l}{\textit{Stage 2 — Retrieval SFT (1 epoch)}} \\
\quad Learning rate    & $2{\times}10^{-5}$ & $1{\times}10^{-4}$ \\
\quad LR scheduler     & Linear & Linear \\
\midrule
\multicolumn{3}{l}{\textit{Shared}} \\
\quad Warmup ratio     & \multicolumn{2}{c}{0.03} \\
\quad Effective batch  & \multicolumn{2}{c}{1024} \\
\bottomrule
\end{tabular}
\caption{Training hyperparameters for both stages.}
\label{tab:hparams-s1}
\end{table}

\subsection*{LoRA Configuration}

Adapters are applied to all linear projection layers (\texttt{q\_proj}, \texttt{k\_proj}, \texttt{v\_proj}, \texttt{o\_proj}, \texttt{gate\_proj}, \texttt{up\_proj}, \texttt{down\_proj}) with rank $r{=}64$, $\alpha{=}128$, and zero dropout.
The full embedding layer (\texttt{embed\_tokens}) is kept fully trainable in all LoRA runs; weight tying between \texttt{embed\_tokens} and \texttt{lm\_head} is preserved.
LoRA Stage~2 runs initialise from the merged Stage~1 LoRA checkpoint.


\section{Realistic-Query Stage~2 Training}
\label{sec:appendix-realistic-training}

To test whether the generalization collapse is caused by query distribution shift or a deeper property of the Stage~2 training objective, we retrained the TG (Gemma3-4B) configuration using RRB-style queries in place of the standard verbose ToolBench queries.

\paragraph{Training data.}
We used ToolSense to generate realistic queries for all $\sim$47k tools in the ToolBench catalog.
Since RRB queries can have multiple correct tools (Medium and Hard tiers require 2--3 and 4+ tools respectively), each generated sample $(q, [t_1, \ldots, t_n])$ is expanded into $n$ individual training pairs $(q, t_1), \ldots, (q, t_n)$, yielding 284,567 training samples in total.
The 500-query RRB evaluation set is disjoint from this training corpus.
All other Stage~2 hyperparameters are identical to the standard TG run (see Appendix~\ref{sec:appendix-hyperparams}).

\paragraph{Results.}
Table~\ref{tab:realistic-training} compares TG trained on standard vs.\ RRB-style Stage~2 data. RRB $R_c@50$ rises by $+$44.0pp (43.8\%$\to$87.8\%), demonstrating that the model has sufficient capacity to handle realistic queries when trained on them.
The cost is a 9.1pp drop on G1 verbose recall (95.7\%$\to$86.6\%), confirming that the two query styles induce complementary but non-overlapping retrieval behaviors and the model specializes to whichever distribution it trains on.
Critically, MCQ accuracy drops from 31.5\% to 26.0\% (near-random), and QA from 50.0\% to 46.6\%, despite the substantial improvement in retrieval.
This demonstrates that Stage~2 SFT erodes parametric tool knowledge regardless of whether training queries are verbose or realistic, implicating the retrieval fine-tuning objective itself rather than the query distribution.

\begin{table*}[h]
\centering
\tiny
\begin{tabular}{lcccccc}
\toprule
Training data & RRB $R_c@50$ & G1 $R_c@50$ & G2 $R_c@50$ & G3 $R_c@50$ & MCQ & QA \\
\midrule
Standard (verbose)    & 43.8\,{\tiny[40.3, 47.7]} & 95.7\,{\tiny[94.2, 97.0]} & 84.5\,{\tiny[81.5, 87.3]} & 84.6\,{\tiny[80.4, 88.5]} & 31.5\,{\tiny[27.6, 35.5]} & 50.0\,{\tiny[45.8, 54.0]} \\
RRB-style (realistic) & 87.8\,{\tiny[85.6, 90.2]} & 86.6\,{\tiny[84.2, 89.1]} & 67.2\,{\tiny[63.6, 70.7]} & 61.8\,{\tiny[54.7, 69.6]} & 26.0\,{\tiny[22.4, 30.0]} & 46.6\,{\tiny[42.4, 50.8]} \\
\bottomrule
\end{tabular}
\caption{TG Gemma3-4B: standard vs.\ RRB-style Stage~2 training (final checkpoint). 95\% bootstrap CIs in brackets.}
\label{tab:realistic-training}
\end{table*}


\begin{table}[t]
\small
\centering
\begin{tabular}{lcc}
\toprule
Stage~1 metric & Domain~A (HR) & Domain~B (Finance) \\
\midrule
MCQ accuracy & 0.50 & 0.52 \\
QA accuracy  & 0.67 & 0.80 \\
\bottomrule
\end{tabular}
\caption{Stage~1 (TG-3FM-LoRA, Gemma3-4B-it) probing accuracy on a closed proprietary
enterprise catalog of 8{,}283 tools never present in any public corpus. Random baselines:
MCQ${=}$25\%, QA${=}$50\%.}
\label{tab:contamination}
\end{table}

\section{Contamination Control: Closed Enterprise Catalog}
\label{sec:appendix-contamination}

ToolBench is derived from the public RapidAPI ecosystem, so open base models may have
encountered its API documentation during pre-training, potentially inflating the
\emph{absolute} Stage~1 knowledge scores. To isolate genuine parametric learning from
pre-training recall, we applied the ToolSense Stage~1 memorization recipe (TG-3FM-LoRA)
to a \emph{closed proprietary} enterprise catalog of 8{,}283 tools that has never appeared
in any public corpus, spanning two internal business domains: Domain~A (HR, 918 tools)
and Domain~B (Finance, 7{,}365 tools). The base model, Gemma3-4B-it, has no possible
pre-training exposure to these tools.

We then evaluated ToolSense-generated MCQ and QA probes over this private catalog
(MCQ: 300/400 items for Domain~A/B; QA: 200/400 items for Domain~A/B; random baselines
MCQ${=}$25\%, QA${=}$50\%). Table~\ref{tab:contamination} reports Stage~1 accuracy.

Both metrics are substantially above random chance on tools with zero possibility of
pre-training exposure, demonstrating that Stage~1 memorization induces \emph{genuine}
parametric tool learning rather than surfacing pre-trained memory. This confirms that
(i)~the absolute Stage~1 scores on ToolBench are not an artifact of contamination, and
(ii)~the ToolSense framework transfers cleanly to closed, proprietary catalogs. In any
case, our central diagnostic claims concern the \emph{relative} Stage~1$\rightarrow$Stage~2
forgetting rather than absolute Stage~1 magnitudes, and are therefore robust to any
residual contamination in ToolBench.


\section{Internalization Score: Rationale and Degenerate-Case Handling}
\label{sec:appendix-is-rationale}

\paragraph{What IS measures and what it does not.}
IS is a \emph{conditional} metric: it asks, ``the tool knowledge the model has already acquired (as measured by $R_c@k$), how much of it has been internalized into the weights rather than delegated to the trie?''
It is \emph{not} a measure of how much the model has learned overall, as that role belongs to $R_c@k$ itself.
Consider the case where $R_c@50 = 0.10$ and $R_f@50 = 0.10$: recall scores are low, indicating the model has learned only about small fraction of tools; but whatever it did learn can be generated freely without constraint, giving IS $= 1.0$.
This is not a failure of the metric, instead the correct reading would be: \emph{whatever small amount was learned is fully internalized}.
The learning deficit is plainly visible in $R_c$; IS tells us about the quality of that learning, not its quantity.
For this reason we always report $R_f$, $R_c$, and IS together (see Tables~\ref{tab:is-full-4b} and~\ref{tab:is-full-scale}) so the reader can distinguish low-recall-but-internalized from high-recall-but-trie-dependent regimes at a glance.

\paragraph{Degenerate case: $R_c@k = 0$.}
When constrained recall is exactly zero the ratio is undefined; we define IS $= 0$ in this case (reflected in Eq.~\ref{eq:is}).
Operationally, $R_c = 0$ means the model cannot recall any correct tool even with full trie guidance, so there is nothing to internalize.
Assigning IS $= 0$ rather than leaving it undefined prevents spurious high-IS readings caused by a model that is uniformly broken.

\paragraph{Relationship to a difference-based formulation.}
An alternative design would use $\Delta @k = R_f@k - R_c@k$.
This directly measures the absolute gap but conflates two distinct phenomena: a model with $R_c = 0.90$, $R_f = 0.80$ (IS $= 0.89$, $\Delta = -0.10$) and one with $R_c = 0.10$, $R_f = 0.00$ (IS $= 0$, $\Delta = -0.10$) would produce the same difference score despite having very different failure modes.
The ratio, always paired with $R_c$, is more interpretable: IS $= 0.89$ at $R_c = 0.90$ means a high-recall model is mostly but not fully internalized; IS $= 0$ at $R_c = 0.10$ means a low-recall model is entirely trie-dependent.


\section{Internalization Score: Full Results with Confidence Intervals}
\label{sec:appendix-is-full}

Tables~\ref{tab:is-full-4b} and~\ref{tab:is-full-scale} report IS@50 for all
evaluated configurations, including 95\% bootstrap confidence intervals on all four
evaluation splits.
CIs for IS are derived via the delta method from the bootstrap CIs of $R_f@50$ and
$R_c@50$: letting $\hat{\sigma}_f$ and $\hat{\sigma}_c$ be the half-widths of the
respective 95\% bootstrap intervals divided by 1.96,
$\sigma_{\text{IS}} \approx \text{IS} \cdot \sqrt{(\hat{\sigma}_f/R_f)^2 +
(\hat{\sigma}_c/R_c)^2}$, and the CI is
$[\text{IS} - 1.96\,\sigma_{\text{IS}},\; \text{IS} + 1.96\,\sigma_{\text{IS}}]$,
clipped to $[0,1]$.

\begin{table*}[h]
  \centering
  \small
  \begin{tabular}{lcccc}
    \toprule
    Config & G1 IS@50 & G2 IS@50 & G3 IS@50 & RRB IS@50 \\
    \midrule
    TG Gemma3-4B & 0.75\,{\tiny[0.72,0.79]} & 0.87\,{\tiny[0.82,0.93]} & 1.00\,{\tiny[0.93,1.00]} & 0.85\,{\tiny[0.74,0.96]} \\
    TG (LoRA) Gemma3-4B & 0.55\,{\tiny[0.50,0.59]} & 0.70\,{\tiny[0.64,0.75]} & 0.95\,{\tiny[0.87,1.00]} & 0.80\,{\tiny[0.68,0.93]} \\
    TG-SP Gemma3-4B & 0.70\,{\tiny[0.66,0.74]} & 0.70\,{\tiny[0.64,0.76]} & 0.65\,{\tiny[0.55,0.75]} & 0.75\,{\tiny[0.65,0.85]} \\
    TG-3FM Gemma3-4B & 0.69\,{\tiny[0.65,0.73]} & 0.78\,{\tiny[0.73,0.84]} & 1.00\,{\tiny[0.93,1.00]} & 0.82\,{\tiny[0.71,0.93]} \\
    TG-3FM (LoRA) Gemma3-4B & 0.89\,{\tiny[0.86,0.93]} & 0.92\,{\tiny[0.87,0.98]} & 1.00\,{\tiny[0.94,1.00]} & 0.81\,{\tiny[0.71,0.91]} \\
    \midrule
    TG-H Gemma3-4B & 0.42\,{\tiny[0.38,0.46]} & 0.41\,{\tiny[0.36,0.46]} & 0.62\,{\tiny[0.53,0.70]} & 0.33\,{\tiny[0.24,0.41]} \\
    TG-5FM Gemma3-4B & 0.84\,{\tiny[0.80,0.87]} & 0.78\,{\tiny[0.72,0.84]} & 0.52\,{\tiny[0.42,0.61]} & 0.64\,{\tiny[0.53,0.76]} \\
    TG-5FM (LoRA) Gemma3-4B & 0.66\,{\tiny[0.62,0.70]} & 0.79\,{\tiny[0.73,0.85]} & 0.61\,{\tiny[0.50,0.72]} & 0.79\,{\tiny[0.65,0.92]} \\
    \bottomrule
  \end{tabular}
  \caption{IS@50 with 95\% confidence intervals for training Gemma3-4B (Stage~2).
           Values in brackets are 95\% CIs.}
  \label{tab:is-full-4b}
\end{table*}

\begin{table*}[h]
  \centering
  \small
  \begin{tabular}{lcccc}
    \toprule
    Config & G1 IS@50 & G2 IS@50 & G3 IS@50 & RRB IS@50 \\
    \midrule
    TG-H Qwen3.5-4B & 0.57\,{\tiny[0.53,0.61]} & 0.49\,{\tiny[0.43,0.54]} & 0.64\,{\tiny[0.55,0.73]} & 0.28\,{\tiny[0.21,0.35]} \\
    TG-3FM Qwen3.5-4B & 0.79\,{\tiny[0.75,0.83]} & 0.78\,{\tiny[0.72,0.85]} & 1.00\,{\tiny[0.91,1.00]} & 0.81\,{\tiny[0.73,0.89]} \\
    TG-3FM (LoRA) Qwen3.5-4B & 0.68\,{\tiny[0.64,0.72]} & 0.77\,{\tiny[0.71,0.83]} & 0.75\,{\tiny[0.66,0.85]} & 0.82\,{\tiny[0.73,0.92]} \\
    TG-SP Qwen3.5-4B     & 0.90\,{\tiny[0.86,0.93]} & 0.97\,{\tiny[0.91,1.00]} & 1.00\,{\tiny[0.92,1.00]} & 0.98\,{\tiny[0.89,1.00]} \\
    TG-SP (LoRA) Qwen3.5-4B & 0.94\,{\tiny[0.92,0.97]} & 0.91\,{\tiny[0.85,0.96]} & 1.00\,{\tiny[0.92,1.00]} & 0.99\,{\tiny[0.89,1.00]} \\
    \midrule
    TG-3FM (LoRA) Gemma3-12B & 0.76\,{\tiny[0.72,0.80]} & 0.85\,{\tiny[0.80,0.90]} & 0.96\,{\tiny[0.88,1.00]} & 0.86\,{\tiny[0.76,0.96]} \\
    TG-SP Gemma3-12B     & 0.66\,{\tiny[0.62,0.70]} & 0.85\,{\tiny[0.80,0.90]} & 0.65\,{\tiny[0.54,0.75]} & 0.84\,{\tiny[0.75,0.93]} \\
    \bottomrule
  \end{tabular}
  \caption{IS@50 with 95\% confidence intervals for training Qwen3.5-4B and Gemma3-12B (Stage~2).}
  \label{tab:is-full-scale}
\end{table*}

\begin{figure*}[h]
  \centering
  \includegraphics[width=\linewidth]{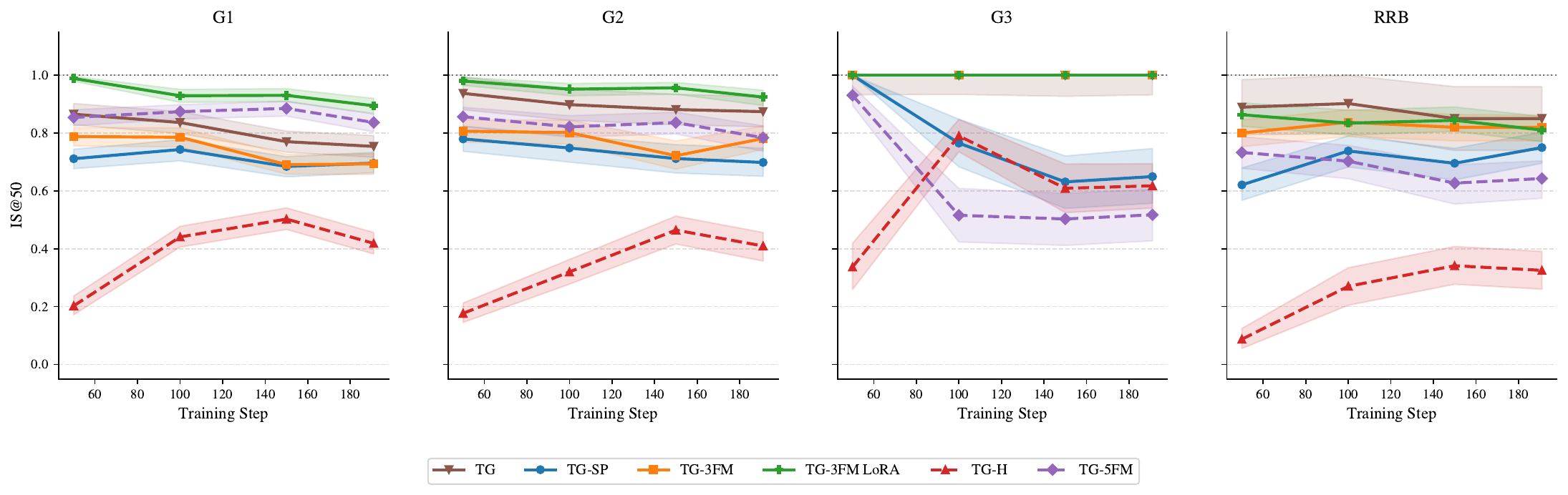}
  \caption{IS@50 over Stage~2 training for Gemma3-4B across all four evaluation splits.
           Shaded bands are 95\% bootstrap CIs.}
  \label{fig:is-appendix-4b}
\end{figure*}

\begin{figure*}[h]
  \centering
  \includegraphics[width=\linewidth]{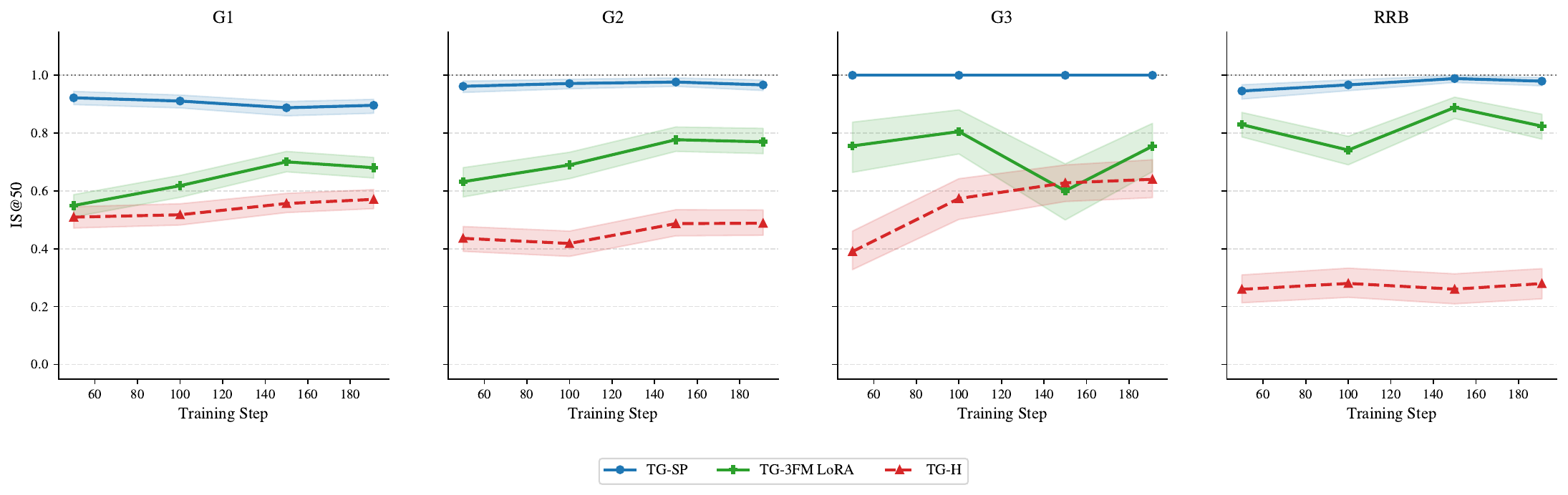}
  \caption{IS@50 over Stage~2 training for Qwen3.5-4B across all four evaluation splits.
           Shaded bands are 95\% bootstrap CIs.}
  \label{fig:is-appendix-scale}
\end{figure*}

The horizontal rule within each table separates flat-token configurations (top) from
hierarchical-token configurations (bottom).
Figures~\ref{fig:is-appendix-4b} and~\ref{fig:is-appendix-scale} show the IS@50 training dynamics
for different configurations across all splits, complementing the main-body Figure~\ref{fig:is-evolution}.

Several patterns are consistent across architectures and sizes:
(i)~flat configurations typically achieve IS${>}0.65$ on both G1 and RRB, with LoRA
variants systematically higher ($\geq 0.80$);
(ii)~hierarchical configurations show a pronounced penalty on RRB---TG-H drops
to IS$=$0.33 on RRB vs.\ 0.42 on G1 for Gemma3-4B, and this pattern holds for Qwen3.5-4B
(0.28 on RRB);
(iii)~G3 IS values are frequently 1.00 or near-1.00 for flat configurations, suggesting
that out-of-distribution single-API queries are handled without trie support;
(iv)~CI widths are wider on RRB than on ToolBench evaluation splits (G1/G2/G3), reflecting the models cause greater instability in consistently producing the tool tokens in organic manner.


\section{Cross-Architecture Probing Results}
\label{sec:appendix-probes-scale}

Table~\ref{tab:probes-scale} reports MCQ and QA probing accuracy for Qwen3.5-4B and
Gemma3-12B, complementing the Gemma3-4B results in Table~\ref{tab:probes}.
Table~\ref{tab:correlation-variants} lists all 14 model variants included in the
Stage~1 MCQ vs.\ Stage~2 RRB correlation analysis (Figure~\ref{fig:correlation}).

\begin{table}[t]
  \centering
  \small
  \begin{tabular}{ll}
    \toprule
    Model family & Configuration \\
    \midrule
    \multirow{7}{*}{Gemma3-4B}
      & TG \\
      & TG-SP \\
      & TG-3FM \\
      & TG-3FM~(LoRA) \\
      & TG-H \\
      & TG-5FM \\
      & TG-5FM~(LoRA) \\
    \midrule
    \multirow{5}{*}{Qwen3.5-4B}
      & TG-3FM \\
      & TG-3FM~(LoRA) \\
      & TG-H \\
      & TG-SP \\
      & TG-SP~(LoRA) \\
    \midrule
    \multirow{2}{*}{Gemma3-12B}
      & TG-3FM~(LoRA) \\
      & TG-SP \\
    \bottomrule
  \end{tabular}
  \caption{All 14 model variants included in the Stage~1 MCQ vs.\ Stage~2 RRB $R_c@50$ correlation plot (Figure~\ref{fig:correlation}).}
  \label{tab:correlation-variants}
\end{table}

The cross-architecture results reveal a consistent pattern: larger and more capable base
models enter Stage~2 with substantially stronger parametric tool knowledge (Stage~1 MCQ
of 73--75\% for Qwen3.5-4B and Gemma3-12B, vs.\ 23--55\% for Gemma3-4B) and largely
\emph{retain} that knowledge after retrieval fine-tuning.
In contrast to Gemma3-4B FFT variants which suffer severe MCQ drops (e.g., TG drops from
55.4\% to 31.4\%), the Qwen3.5-4B and Gemma3-12B configurations maintain or even
slightly improve MCQ scores --- confirming that knowledge destruction is more pronounced
in smaller models fine-tuned with full parameter updates.
Hierarchical tokenization (TG-H) remains the most destructive configuration across
architectures: Qwen3.5-4B TG-H drops from 39.7\% to 31.7\% MCQ, reinforcing the
earlier finding that collapsing tool tokens into path-structured representations
impedes semantic retention regardless of model scale.
LoRA consistently provides the best knowledge preservation for Qwen3.5-4B
(TG-3FM LoRA retains 74.2\% MCQ through Stage~2), confirming its role as the safest
fine-tuning strategy when knowledge retention is required.

\begin{table}[t]
  \centering
  \resizebox{\columnwidth}{!}{%
  \begin{tabular}{llcccc}
    \toprule
    Model & Config & \multicolumn{2}{c}{MCQ (S1$\rightarrow$S2)} & \multicolumn{2}{c}{QA (S1$\rightarrow$S2)} \\
    \cmidrule(lr){3-4} \cmidrule(lr){5-6}
          &        & S1 & S2 & S1 & S2 \\
    \midrule
    \multirow{5}{*}{Qwen3.5-4B}
      & TG-3FM        & \textbf{74.0\,{\small[70.4,77.8]}} & 72.8\,{\small[69.2,76.6]} & 53.6\,{\small[49.2,58.0]} & 54.2\,{\small[49.8,58.8]} \\
      & TG-3FM (LoRA) & 73.8\,{\small[69.6,77.8]} & \textbf{74.2\,{\small[70.6,78.0]}} & \textbf{64.2\,{\small[60.2,68.2]}} & 54.4\,{\small[50.0,58.8]} \\
      & TG-H          & 39.7\,{\small[35.3,44.2]} & 31.7\,{\small[27.6,36.1]} & 49.6\,{\small[44.8,54.0]} & 50.8\,{\small[46.6,55.2]} \\
      & TG-SP         & 63.1\,{\small[58.9,67.3]} & 61.7\,{\small[57.5,65.7]} & 51.6\,{\small[47.8,56.2]} & 48.0\,{\small[43.8,52.4]} \\
      & TG-SP (LoRA)  & 64.9\,{\small[60.5,69.2]} & 49.0\,{\small[44.6,53.8]} & 51.4\,{\small[47.0,55.6]} & \textbf{51.6\,{\small[47.8,56.2]}} \\
    \midrule
    \multirow{2}{*}{Gemma3-12B}
      & TG-3FM (LoRA) & \textbf{75.0\,{\small[71.0,78.8]}} & \textbf{76.4\,{\small[72.6,80.0]}} & 62.4\,{\small[57.8,66.8]} & 65.6\,{\small[61.4,69.6]} \\
      & TG-SP         & 60.5\,{\small[56.1,65.1]} & 66.3\,{\small[62.1,70.4]} & \textbf{66.4\,{\small[62.0,70.6]}} & \textbf{71.2\,{\small[67.0,75.2]}} \\
    \bottomrule
  \end{tabular}}
  \caption{MCQ and QA probing accuracy for Qwen3.5-4B and Gemma3-12B. Random baselines: MCQ=25\%, QA=50\%. 95\% CI in brackets.}
  \label{tab:probes-scale}
\end{table}


\section{Virtual Token Embedding Drift Analysis}
\label{sec:appendix-emb-drift}

To understand what Stage~2 SFT modifies mechanistically, we measure the \emph{relative L2 drift} of virtual token embeddings between Stage~1 and Stage~2 checkpoints.
For a token index $i$, define:
\begin{equation}
  d_\mathrm{rel}(i) = \frac{\|\mathbf{E}_{S2}[i] - \mathbf{E}_{S1}[i]\|_2}{\|\mathbf{E}_{S1}[i]\|_2}
  \label{eq:emb-drift-app}
\end{equation}
where $\mathbf{E}_{S1}, \mathbf{E}_{S2} \in \mathbb{R}^{V \times d}$ are the embedding matrices at Stages~1 and~2 ($d{=}3840$ for Gemma-3-4B-IT). We compare virtual tokens against 5{,}000 randomly sampled base-vocabulary tokens using a one-sided Mann-Whitney U test. Results across all seven Gemma3-4B configurations are shown in Figure~\ref{fig:emb-drift}.

\paragraph{Finding 1: Stage~2 selectively repositions virtual tokens (all configs, $p < 2{\times}10^{-308}$).}
Drift ratios range from $1.9\times$ to $22.9\times$ depending on training regime. This rules out uniform global drift: if Stage~2 were broadly fine-tuning all weights, virtual and regular tokens would experience similar relative displacement.
In FFT models ($2$--$3\times$ ratio), virtual tokens appear in \emph{every} training example as generation targets while regular tokens appear sparsely in user queries; by the law of large numbers over mini-batches, virtual tokens accumulate proportionally larger gradient updates~\citep{li2021prefix,lester2021prompt}.

\paragraph{Finding 2: LoRA creates extreme isolation ($22.9\times$).}
Under LoRA~\citep{hu2022lora}, adapters are attached to linear layers while the full embedding layer remains trainable.
Despite this, base vocabulary tokens drift by only $D_\mathrm{rel}(\mathrm{RAND}){=}0.0015$ — effectively numerical noise — because they appear solely in input queries and their gradients must back-propagate through the frozen backbone layers before reaching the embedding table, arriving with negligible magnitude.
Virtual tokens, as generation targets in every Stage~2 training example, receive direct strong gradient from the output cross-entropy loss.
This input/output asymmetry, amplified by backbone freezing, is responsible for the $22.9\times$ drift ratio.

\paragraph{Finding 3: Hierarchical tokens occupy a near-isotropic geometric regime.}
Flat tokens have mean pairwise cosine similarity $\mathrm{cosim}_{S1}{\approx}0.37$, a signature of embedding anisotropy produced when all tool tokens share similar Stage~1 generation objectives, pushing them into a common semantic cone~\citep{gao2019degeneration,ethayarajh2019contextual,timkey2021rogue}.
Hierarchical tokens have $\mathrm{cosim}_{S1}{\approx}0.038$, approaching the isotropic baseline expected for randomly initialised unit vectors. For Gemma-3's weight-tied architecture ($\mathbf{W}_\mathrm{head}{=}\mathbf{E}$), the trie-routing logit margin is $l_i - l_j = (\mathbf{E}[i] - \mathbf{E}[j])^\top \mathbf{h}$; near-orthogonal embeddings maximise this margin, so hierarchical tokens are better conditioned for constrained decoding from the start~\citep{decao2021genre}.
Consequently, hierarchical tokens require $7$--$24\times$ less Stage~2 repositioning (TG-3FM flat: $d_\mathrm{rel}{=}0.024$; TG-5FM hier: $d_\mathrm{rel}{=}0.003$).

\paragraph{Finding 4: No representation collapse ($|\Delta\mathrm{cosim}| < 0.002$).}
Despite targeted repositioning, pairwise cosine similarity of virtual tokens is essentially unchanged by Stage~2 across all configurations. This rules out the hypothesis that Stage~2 collapses virtual tokens into undifferentiated routing codes. Each token retains its Stage-1-acquired geometric identity; what changes is absolute position (drift), not relative configuration (cosim). The trie objective structurally prevents collapse: if all tokens became identical, constrained beam search could not route to specific tools.

\paragraph{Connection to knowledge-retrieval dissociation.}
Despite the targeted and statistically significant Stage~2 updates, MCQ/QA accuracy on tool knowledge degrades, confirming that Stage~2 optimizes virtual token positions as trie-routing codes rather than encoding semantic tool knowledge.
Even hierarchical tokens, geometrically superior in every metric and requiring minimal Stage~2 adaptation, fail MCQ probes, closing the confound that poor token geometry explains the dissociation.
This is consistent with causal analyses locating factual recall in MLP layers of mid-to-upper transformer blocks~\citep{meng2022rome,dai2022knowledge,petroni2019lama}: Stage~2 targets the input embedding layer, leaving factual recall pathways untouched~\citep{sun2024prefixtheory}.

\section{Benchmark Generation Details}
\label{sec:appendix-benchmark-gen}

All three ToolSense benchmarks (RRB, MCQ, QA) were generated using \textbf{claude-4.5-sonnet} for both generation and judging calls, accessed via a LiteLLM proxy.
Hard-negative retrieval for the RRB pipeline used OpenAI \texttt{text-embedding-3-large} embeddings stored in a ChromaDB vector database.
Judge calls were made at temperature~$=0.0$; generation calls used the model's default sampling temperature.

\subsection{Realistic Retrieval Benchmark (RRB) Generation Pipeline}
\label{sec:appendix-rrb-gen}

\paragraph{Seed selection.}
Anchor tools are sampled from the ToolBench catalog using stratified sampling across service domains (parsed from the \texttt{Service\&\&Method} name format), with proportional allocation and a minimum of one seed per domain.

\paragraph{Hard-negative pool construction.}
For each anchor tool~$\tau$, 13 hard negatives are retrieved from the full catalog via cosine similarity over \texttt{text-embedding-3-large} embeddings of tool descriptions.
The candidate pool is $\mathcal{P}(\tau) = \{\tau\} \cup \{\text{13 hard negatives}\}$, giving 14 total candidates.

\paragraph{Generation and validation pipeline.}
Generation is implemented as a LangGraph \texttt{StateGraph} with three tier subgraphs (easy, medium, hard) running in parallel via fan-out/fan-in.
Each subgraph follows: \emph{generate} $\to$ \emph{programmatic filter} $\to$ \emph{LLM judge} $\to$ (accept or retry with feedback).
Up to three retries are allowed per sample; on retry, the judge's rejection reason is injected into the next generation prompt.
The programmatic filter checks that all answer tool names appear verbatim in the candidate pool, preventing hallucinated tool names.

\paragraph{Tier specifications.}
Easy tier queries must resolve to exactly 1 tool, remain under 20~words, and be phrased in business language.
Medium tier queries must be genuinely ambiguous across 2--3 tools (overlapping functional domains), under 25~words.
Hard tier queries describe a high-level business goal spanning 4+ tools, in 1--3 sentences.
Default sample targets per seed: 3 (easy), 3 (medium), 2 (hard).

\subsection{MCQ and QA Probing Benchmark Generation}
\label{sec:appendix-probe-gen}

Both benchmarks follow the same two-step generate-then-judge pattern: (1)~a generation call produces a candidate item; (2)~a judge call at temperature~$=0.0$ accepts or rejects it.
Items failing the judge or flagged by the generator as unanswerable (\texttt{skip=true}) are discarded.

\paragraph{MCQ.}
For each tool, the generator produces a 4-way multiple-choice question: one correct answer and three plausible-but-wrong distractors, all as short phrases (2--8 words).
Questions must use \textit{``this tool''} as a placeholder and test a specific, verifiable property.
The judge verifies all five criteria: specificity, correctness, distractor plausibility, option distinctiveness, and placeholder usage.
Final dataset: 496 items.

\paragraph{QA.}
For each tool, the generator produces one binary yes/no question with a pre-specified target answer (alternated across tools for label balance).
Questions must use \textit{``this tool''} as a placeholder and test a verifiable property (modality, domain, input/output type, etc.).
The judge verifies four criteria: specificity, correctness, placeholder usage, and verifiability.
Final dataset: 500 items.

\section{System Prompts}
\label{sec:appendix-prompts}

We report the exact user-turn prompts used for experiments conducted in this study.
Placeholders in \texttt{\{braces\}} are filled at runtime with the corresponding value.

\subsection{Memorization Format Prompts}
\label{sec:appendix-prompts-memo}

The \texttt{desc$\to$tok} format (Figure~\ref{fig:prompt-desc-tok}), used in all
configurations, asks the model to predict the virtual token from the tool's JSON
description.
For flat tokens the expected output is a single atomic identifier
(\texttt{<<API\&\&ENDPOINT>>}); for hierarchical tokens it is the joint two-token
sequence (\texttt{<<API>><<ENDPOINT>>}) generated autoregressively.

The \texttt{tok$\to$desc} format (Figure~\ref{fig:prompt-tok-desc}), used in TG-3FM
and TG-5FM, reverses the mapping: the model receives the virtual token and is asked to generate the plain-language tool description.
This bidirectional objective makes the association more resistant to overwriting
during Stage~2.

The \texttt{desc$\to$api\_tok} format (Figure~\ref{fig:prompt-desc-api-tok}), used
only in TG-5FM, restricts the target to the API-level token alone, providing
an additional supervised signal that trains the first hierarchical generation step
independently.

The \texttt{desc+api\_tok$\to$endpoint\_tok} format
(Figure~\ref{fig:prompt-desc-svc-endpoint}), also TG-5FM only, gives the model both
the tool description and the API-level token and asks it to predict the endpoint-level
token, training the second hierarchical step independently.

The MCTS format (Figure~\ref{fig:prompt-mcts}), used in TG-3FM and TG-5FM, presents
a discriminative selection task: the model must choose the correct virtual token from
$K+1$ hard-negative candidates retrieved by embedding similarity.
See Appendix~\ref{sec:appendix-mcts} for construction details.

\begin{figure}[h]
  \begin{tcolorbox}[promptstyle]

You will be provided with the information about the tool in the JSON format. Your task is to predict the tool that corresponds to the given tool information. You only need to predict the tool. Please don't include any additional text in your response. The tool information is provided below:\par\smallskip%
\{tool\_info\}
  \end{tcolorbox}
  \caption{\texttt{desc$\to$tok} prompt (all configurations).}
  \label{fig:prompt-desc-tok}
\end{figure}

\begin{figure}[h]
  \begin{tcolorbox}[promptstyle]

You will be provided with a tool representation tokens. Your task is to describe what this tool does in plain language. Please don't include any additional text in your response other than the description. The tool representation is provided below:\par\smallskip%
\{tool\_token\}
  \end{tcolorbox}
  \caption{\texttt{tok$\to$desc} prompt (TG-3FM, TG-5FM).}
  \label{fig:prompt-tok-desc}
\end{figure}

\begin{figure}[h]
  \begin{tcolorbox}[promptstyle]

You will be provided with the information about the tool in the JSON format. Your task is to identify the API this tool belongs to and produce its corresponding API token. Only respond with the API token and nothing else. The tool information is provided below:\par\smallskip%
\{tool\_info\}
  \end{tcolorbox}
  \caption{\texttt{desc$\to$api\_tok} prompt (TG-5FM only).}
  \label{fig:prompt-desc-api-tok}
\end{figure}

\begin{figure}[h]
  \begin{tcolorbox}[promptstyle]

You will be provided with the information about the tool in the JSON format along with its API. Your task is to identify the Endpoint this tool belongs to within that API and produce its corresponding Endpoint token. Only respond with the Endpoint token and nothing else. The tool information and API are provided below:\par\smallskip%
Tool Information: \{tool\_info\}\par\smallskip%
API: \{api\_token\}
  \end{tcolorbox}
  \caption{\texttt{desc+api\_tok$\to$endpoint\_tok} prompt (TG-5FM only).}
  \label{fig:prompt-desc-svc-endpoint}
\end{figure}

\begin{figure}[h]
  \begin{tcolorbox}[promptstyle]

You will be provided with the description of a tool along with line-separated multiple choice options for the tool tokens. Your task is to select the correct tool token that corresponds to the given tool description. Please respond with only the correct tool token without any additional text. The tool description and options are provided below:\par\smallskip%
{[START OF TOOL DESCRIPTION]}\\%
\{tool\_description\}\\%
{[END OF TOOL DESCRIPTION]}\par\smallskip%
{[START OF OPTIONS]}\\%
\{tool\_options\}\\%
{[END OF OPTIONS]}\par\smallskip%
From the above new line-separated options, select the correct tool. Only respond with the tool representation (sequence of tokens that represent tool) and nothing else.
  \end{tcolorbox}
  \caption{MCTS discriminative prompt (TG-3FM, TG-5FM).}
  \label{fig:prompt-mcts}
\end{figure}

\subsection{Retrieval Training Prompt}
\label{sec:appendix-prompts-retrieval}

Figure~\ref{fig:prompt-retrieval} shows the Stage~2 prompt shared by all configurations
that use a system prompt (TG-SP, TG-3FM, TG-H, TG-5FM).
The model receives a natural-language user query and must predict the corresponding
virtual tool token.
The no-prompt baseline TG omits this prompt entirely; all other configurations use it
consistently across Stage~2 training and inference.

\begin{tcolorbox}[promptstyle]
\tiny
You will be provided with the user query. Your task is to predict the tool that can best fulfill the user's request. You only need to predict the tool token. Please don't include any additional text in your response. The user query is provided below:\par\smallskip%
\{query\}
\end{tcolorbox}
\captionof{figure}{Stage~2 retrieval training and inference prompt.}
\label{fig:prompt-retrieval}

\subsection{Benchmark Generation Prompts}
\label{sec:appendix-gen-prompts}

All benchmark generation prompts were used with \textbf{claude-4.5-sonnet} via a LiteLLM proxy.
Figures~\ref{fig:prompt-qa-gen} and~\ref{fig:prompt-qa-judge} show the generation and judge prompts for the QA probing benchmark; Figures~\ref{fig:prompt-mcq-gen} and~\ref{fig:prompt-mcq-judge} show the corresponding prompts for MCQ.
All judge prompts run at temperature~$=0.0$.
Figure~\ref{fig:prompt-rrb-easy} shows the full RRB easy-tier generation prompt (Jinja2 template); the medium- and hard-tier prompts share the same structure with tier-specific rules substituted in.
Figure~\ref{fig:prompt-rrb-validator} shows the shared RRB judge prompt, parameterised by the \texttt{\{\{complexity\}\}} variable.


\begin{tcolorbox}[promptstyle]
\tiny
You are validating a yes/no Q\&A entry for an AI tool probing benchmark.

Tool description:\\
\{description\}

Generated entry:\\
\quad Question~: \{question\}\\
\quad Answer~~~: \{answer\}

Check ALL of the following:\\
1. The question is specific to THIS tool --- not generically answerable for any API.\\
2. The answer (``\{answer\}'') is directly and unambiguously supported by the description.\\
3. The question uses ``this tool'' as a placeholder --- the actual tool name does not appear.\\
4. The question tests a verifiable property (domain, capability, input/output type, format, etc.).

Set accept=true only if ALL four checks pass. Otherwise set accept=false and state which check failed.

\{format\_instructions\}
\end{tcolorbox}
\captionof{figure}{QA probing benchmark: judge prompt (temperature~$=0.0$).}
\label{fig:prompt-qa-judge}

\begin{tcolorbox}[promptstyle]
\small
You are building a factual probing benchmark for AI tool retrieval research.

Below is the description of an API or software tool:\\
\{description\}

Your task: generate one yes/no question that tests knowledge of a specific, verifiable functionality or capability of this tool. The correct answer MUST be ``\{target\_answer\}''.

Rules:\\
1. The answer to your question must be exactly ``\{target\_answer\}'' based on the description.\\
2. Use ``this tool'' in the question --- never include the actual tool name or service name.\\
\quad (The model will see only a token at inference time, so the name must not be a hint.)\\
3. The question must be specific to THIS tool. ``Does this tool provide an API?'' is too generic.\\
\quad Good examples for Yes: ``Does this tool process image inputs?'', ``Is this tool designed for financial data?''\\
\quad Good examples for No: ``Does this tool support voice/audio input?'', ``Does this tool return results in XML?''\\
\quad For No questions: ask about a plausible capability or data type that the description does NOT mention\\
\quad (e.g.\ a different modality, format, or domain that a user might expect but this tool doesn't handle).\\
4. Set skip=true if you cannot form a specific, unambiguous question with the required answer.

\{format\_instructions\}
\end{tcolorbox}
\captionof{figure}{QA probing benchmark: generation prompt.}
\label{fig:prompt-qa-gen}

\begin{tcolorbox}[promptstyle]
\small
You are building a multiple-choice probing benchmark for AI tool retrieval research.

Below is the description of an API or software tool:\\
\{description\}

Your task: generate one specific factual question about this tool's properties and provide one correct short answer plus three wrong-but-plausible alternatives.

Rules:\\
1. Use ``this tool'' in the question --- never include the actual tool name or service name.\\
\quad (The model will see only a virtual token at inference time, so the name must not be a hint.)\\
2. The question must be specific to THIS tool --- not answerable for a generic API.\\
\quad Good topics: primary output type, domain/industry, key input, core capability, supported format.\\
\quad Bad question: ``Does this tool provide an API?'' (too generic).\\
3. Each answer option must be a short phrase (2--8 words), not a full sentence.\\
4. The correct\_answer must be directly supported by the description above.\\
5. The three wrong answers must be plausible alternatives --- the kind of answer a user might\\
\quad expect from a tool in the same domain --- but clearly incorrect for THIS specific tool.\\
6. All four options (correct + wrong) must be meaningfully distinct from each other.\\
7. Set skip=true if you cannot form a specific, unambiguous factual question from this description.

\{format\_instructions\}
\end{tcolorbox}
\captionof{figure}{MCQ probing benchmark: generation prompt.}
\label{fig:prompt-mcq-gen}

\begin{tcolorbox}[promptstyle]
\small
You are validating a multiple-choice question entry for an AI tool probing benchmark.

Tool description:\\
\{description\}

Generated entry:\\
\quad Question\hspace{6pt}: \{question\}\\
\quad Correct answer: \{correct\_answer\}\\
\quad Wrong answers~: \{wrong\_answers\}

Check ALL of the following:\\
1. The question is specific to THIS tool --- not generically answerable for any API.\\
2. The correct answer is directly and unambiguously supported by the description.\\
3. Each of the three wrong answers is plausible for a tool in the same domain but clearly incorrect for THIS tool.\\
4. All four options are meaningfully distinct from each other.\\
5. The question uses ``this tool'' as a placeholder --- the actual tool name does not appear.

Set accept=true only if ALL five checks pass. Otherwise set accept=false and state which check failed.

\{format\_instructions\}
\end{tcolorbox}
\captionof{figure}{MCQ probing benchmark: judge prompt (temperature~$=0.0$).}
\label{fig:prompt-mcq-judge}

\begin{tcolorbox}[promptstyle]
\footnotesize
You are a data generation expert creating an evaluation benchmark for a tool retrieval model.\\
Your task is to generate \{\{n\_samples\}\} evaluation samples, each consisting of a concise enterprise user query that points to EXACTLY ONE specific tool.

{[START OF TASK DESCRIPTION]}\\
The goal is to simulate how real enterprise users phrase requests --- NOT how developers write API calls.\\
Unlike typical benchmark queries (which are overly specific and technical), your queries should be:\\
- Concise: 1--2 sentences, ideally under 20 words\\
- Written in natural business language (NOT using API names, technical identifiers, or schema terms)\\
- Realistic: something a business analyst, manager, or power user would actually type

Each sample must contain:\\
1. A query --- concise, business-language question pointing to the target tool\\
2. An answer --- a list containing exactly ONE tool name (the target tool)\\
{[END OF TASK DESCRIPTION]}

{[START OF GENERATION RULES]}\\
1. Queries must be CONCISE --- avoid verbose, over-specified phrasing\\
2. Use BUSINESS LANGUAGE --- never include API method names, OData syntax, or system-specific technical identifiers in the query\\
3. The query should naturally lead to the target tool without explicitly naming it\\
4. The query should NOT trivially match every other tool in the pool --- it must be specific enough to distinguish the target\\
5. Do NOT use OData query syntax (\$filter, \$select, \$expand, etc.) in queries\\
6. Do NOT hallucinate tool names --- only use names exactly as they appear in the tool list below\\
7. Every query must be distinct --- vary phrasing and angle of attack

GOOD query examples (concise, business language):\\
- ``Show me all open purchase orders''\\
- ``Which customers have overdue invoices?''\\
- ``I need to track employee time-off requests''\\
- ``List products that are low on stock''

BAD query examples (too technical or verbose):\\
- ``Retrieve all PurchaseOrders where Status eq `Open' using the GetPurchaseOrders endpoint''\\
- ``I need to call the employee leave management API to fetch pending time-off approval requests for the current fiscal quarter''\\
{[END OF GENERATION RULES]}

{[START OF TARGET TOOL]}\\
Tool Name: \{\{anchor\_tool\_name\}\}\\
Description: \{\{anchor\_tool\_description\}\}\\
{[END OF TARGET TOOL]}

{[START OF TOOL LIST]}\\
These are all available tools. Your answer must contain exactly one tool name from this list:\\
\{\{candidate\_pool\_str\}\}\\
{[END OF TOOL LIST]}

\{\% if previous\_feedback and previous\_feedback != ``None'' \%\}\\
{[START OF FEEDBACK FROM PREVIOUS ATTEMPT]}\\
\{\{previous\_feedback\}\}\\
Address these issues in your new generation.\\
{[END OF FEEDBACK FROM PREVIOUS ATTEMPT]}\\
\{\% endif \%\}

{[START OF OUTPUT FORMAT]}\\
\{\{format\_instructions\}\}\\
{[END OF OUTPUT FORMAT]}

Generate \{\{n\_samples\}\} easy-tier eval samples. Each query must be concise (under 20 words), in business language, and point unambiguously to exactly one tool.
\end{tcolorbox}
\captionof{figure}{RRB easy-tier generation prompt (Jinja2 template; \texttt{\{\{var\}\}} = template variable, \texttt{\{\%~...\%\}} = control block). The medium-tier prompt is identical except: queries target 2--3 tools with genuinely overlapping functional domains, length limit is 25~words, and the answer is a list of 2--3 tool names. The hard-tier prompt targets 4+ tools spanning a high-level business goal, in 1--3~sentences.}
\label{fig:prompt-rrb-easy}

\begin{tcolorbox}[promptstyle]
\footnotesize
You are a quality validator for an enterprise tool retrieval evaluation benchmark. Given a batch of generated (query, answer) samples, assess whether each sample is suitable as an evaluation example.

{[START OF TASK DESCRIPTION]}\\
Validate each sample and return one verdict per sample (in the same order).\\
For each sample, determine:\\
1. Is the query CONCISE and written in BUSINESS LANGUAGE (not technical API language)?\\
2. Does the query sound like something a real enterprise user would actually ask?\\
3. For \{\{complexity\}\} tier: is the number of answer tools and the ambiguity level appropriate?\\
\quad - easy: query should be specific enough that one tool is the clear answer; the phrasing should not be technical\\
\quad - medium: query should be genuinely ambiguous between 2--3 tools; ambiguity must arise from overlapping domains, not vague phrasing\\
\quad - hard: query should describe a high-level business goal that legitimately spans 4+ tools\\
4. Are all answer tools plausibly relevant to the query?\\
{[END OF TASK DESCRIPTION]}

{[START OF VALIDATION RULES]}\\
A sample PASSES (validation\_result: true) if ALL of the following hold:\\
- The query is CONCISE (1--3 sentences, not verbose or over-specified)\\
- The query uses BUSINESS LANGUAGE --- no API method names, technical identifiers, OData operators (\$filter, \$select, \$expand, \$orderby, etc.)\\
- The query reads naturally --- it sounds like a real enterprise user asking a question, not a developer writing a specification\\
- All answer tools are plausibly relevant to the stated query\\
- The ambiguity level matches the complexity tier

A sample FAILS (validation\_result: false) if ANY of the following hold:\\
- The query is verbose or reads like a technical specification\\
- The query contains API names, OData syntax, or schema property names\\
- The query is too generic to be meaningful (e.g.\ ``get data'', ``manage system entities'')\\
- Answer tools are not justified by the query content\\
- Wrong complexity (e.g.\ an easy query matching 3+ tools, or a hard query where only 1--2 tools fit)\\
{[END OF VALIDATION RULES]}

{[START OF SAMPLES TO VALIDATE]}\\
\{\{samples\_str\}\}\\
{[END OF SAMPLES TO VALIDATE]}

{[START OF OUTPUT FORMAT]}\\
\{\{format\_instructions\}\}\\
{[END OF OUTPUT FORMAT]}

Validate each sample in order and return one verdict per sample.
\end{tcolorbox}
\captionof{figure}{RRB LLM judge prompt, shared across all three tiers (temperature~$=0.0$). The \texttt{\{\{complexity\}\}} variable is instantiated as \texttt{easy}, \texttt{medium}, or \texttt{hard} per tier.}
\label{fig:prompt-rrb-validator}

\section{Artifact Licenses and Intended Use}
\label{sec:appendix-licenses}

\paragraph{Datasets used.}
ToolBench~\citep{qin2023toolllm} provides tool documentation metadata crawled from
RapidAPI's publicly accessible API catalog.
The ToolBench repository and dataset are released under the Apache-2.0 License.
Our use conforms to these terms: we do not redistribute the raw metadata, instead releasing
the generated diagnostic benchmarks (RRB, MCQ, QA) derived from it.

\paragraph{Opensource instruction-tuned models.}
Gemma~3~\citep{google2025gemma3} is released by Google under the Gemma Terms of Use, which
permits research and commercial use subject to the stated conditions.
Qwen3.5~\citep{qwen2025qwen3} is released by Alibaba under the Apache~2.0 License.

\paragraph{Artifacts released by this work.}
The ToolSense benchmark generation framework and the ToolBench diagnostic benchmarks
(RRB, MCQ, QA) are released under the \textbf{Apache~2.0 License}.
The intended use is research on parametric tool-retrieval systems, multi-stage
fine-tuning evaluation, and related NLP benchmarking tasks.
We do not recommend using the benchmarks to make production deployment decisions without
independent validation on the target tool catalog.

\section{Dataset Documentation}
\label{sec:appendix-data-doc}

\paragraph{PII and offensive content.}
The ToolBench tool metadata consists entirely of API endpoint names, descriptions,
and parameter schemas scraped from RapidAPI's public-facing documentation.
This data contains no personally identifiable information (PII), no sensitive categories
(health, finance, biometric, etc.), and no offensive content.
The RRB, MCQ, and QA benchmarks are fully synthetic; all items were generated by an LLM
conditioned on the above API documentation, with no user data or personal information involved.

\paragraph{Dataset statistics.}
Table~\ref{tab:dataset-stats} summarises the benchmarks created in this work.
The ToolBench tool catalog used for training and evaluation contains $\sim$47k tools spanning $\sim$16k API services.

\begin{table}[h]
\centering
\small
\begin{tabular}{llrrr}
\toprule
Benchmark & Split & Easy & Medium & Hard \\
\midrule
RRB & eval & 167 & 167 & 166 \\
\midrule
MCQ & eval & \multicolumn{3}{c}{496 items} \\
QA  & eval & \multicolumn{3}{c}{500 items} \\
\bottomrule
\end{tabular}
\caption{Benchmark item counts.
The RRB evaluation set (500 items total) is stratified across three difficulty tiers.
MCQ and QA are single-split probing benchmarks.}
\label{tab:dataset-stats}
\end{table}

\section{Computational Resources and Software}
\label{sec:appendix-compute}

\paragraph{Hardware.}
All training and evaluation experiments were run on a single node equipped with
8~NVIDIA H200 SXM5 (141\,GB) GPUs.
All model configurations used a single GPU.

\paragraph{Software versions.}
All training code uses the following package versions:
Python~3.13, PyTorch~2.10.0, HuggingFace Transformers~5.5.0, PEFT~0.19.0,
Accelerate~1.11.0, TRL~1.2.0, Datasets~4.7.0, and DeepSpeed~0.18.2.
Benchmark generation used LiteLLM~$\geq$1.60, LangGraph~$\geq$0.2.0,
ChromaDB~$\geq$0.5.0, and Pydantic~$\geq$2.0.0.

\section{Human Annotation Study}
\label{sec:appendix-annotation}

\paragraph{Task instructions.}
Three annotators independently validated a stratified random sample of 100 items per
benchmark following the instructions below.

For \textbf{RRB}: annotators were shown each query and a pool of 14 candidate tools (the
ground-truth tool(s) plus hard-negative neighbors from the generation pipeline).
They were asked: ``Select all tools from the candidate list that you believe are a
correct and relevant match for this query's intent.''
Annotators were instructed to rely solely on the tool names and descriptions provided
and not to use external search.

For \textbf{MCQ}: annotators were shown the full tool description and the four answer
options and asked to select the single correct option (A/B/C/D) based on the description.

For \textbf{QA}: annotators were shown the full tool description and the yes/no question
and asked to answer Yes or No based solely on the description.

\paragraph{Recruitment and compensation.}
Annotators were full-time employees of the authoring organization; no additional compensation beyond their regular employment was provided.

\paragraph{Consent and data protection.}
Annotators were informed of the study's purpose before participating and provided
voluntary consent to have their annotations used for research.
No personal data belonging to the annotators or any third parties was collected or stored.

\paragraph{Ethical review.}
This annotation study involves expert evaluation of AI-generated benchmark items and
does not constitute human subjects research involving personal data collection.

\paragraph{Annotator demographics.}
All three annotators are native or proficient English speakers with deep professional expertise in enterprise-scale tool catalogs and API ecosystems.

\section{Use of AI Assistants}
\label{sec:appendix-ai-assist}

Claude Code with claude-4.5-opus (Anthropic) was used to assist with prototyping parts of the experimental codebase and with refining the writing and presentation of this paper.

\end{document}